\documentclass[journal]{IEEEtran}
\usepackage{cite}

\ifCLASSINFOpdf
   \usepackage[pdftex]{graphicx}
\else
\fi
\usepackage{amsmath}
\usepackage{algorithmic}

\usepackage{array}
\usepackage{booktabs}

\ifCLASSOPTIONcompsoc
  \usepackage[caption=false,font=normalsize,labelfont=sf,textfont=sf]{subfig}
\else
  \usepackage[caption=false,font=footnotesize]{subfig}
\fi
\usepackage{dblfloatfix}
\usepackage{url}

\usepackage{bm}
\usepackage{amssymb}

\usepackage{pgfplots} 
\pgfplotsset{width=0.45\textwidth,compat=1.9}
\usepgfplotslibrary{external}

\newcommand{\R}{\mathbb{R}}

\newcommand{\x}{\bm{x}}
\newcommand{\y}{\bm{y}}

\newcommand{\F}{\bm{\mathcal{F}}}

\newcommand{\diff}{\text{\textup{d}}}
\newcommand{\D}{\mathrm{D}}

\definecolor{myorange}{RGB}{248,132,42}
\definecolor{mygreen}{RGB}{27, 153, 139}
\definecolor{mygray}{RGB}{118, 156, 151}
\definecolor{mylightorange}{RGB}{254, 158, 89}
\definecolor{myred}{RGB}{153, 54, 27}

\newtheorem{proposition}{Proposition}
\newtheorem{definition}{Definition}

\begin{document}
%
\renewcommand{\arraystretch}{1.2} 
\title{3D Solid Spherical Bispectrum CNNs for Biomedical Texture Analysis}
%
%
%

\author{Valentin~Oreiller,
        Vincent~Andrearczyk,
        Julien~Fageot, 
        
        John~O.~Prior,~\IEEEmembership{Senior Member,~IEEE}
        and~Adrien~Depeursinge,~\IEEEmembership{Member,~IEEE}

\thanks{V. Oreiller and A. Depeursinge are with the Institute of Information Systems, University of Applied Sciences Western Switzerland (HES-SO), Sierre, Switzerland and
with the Service of Nuclear Medicine and Molecular Imaging, CHUV, Lausanne, Switzerland. e-mail: valentin.oreiller@hevs.ch, adrien.depeursing@hevs.ch. 
V. Andrearczyk is with the Institute of Information Systems, University of Applied Sciences Western Switzerland (HES-SO), Sierre, Switzerland.
J. Fageot is with Harvard School of Engineering and Applied Sciences, Cambridge, MA, USA.
J. Prior is with the Service of Nuclear Medicine and Molecular Imaging, CHUV, Lausanne, Switzerland.}}
\maketitle

\begin{abstract}
Locally Rotation Invariant (LRI) operators have shown great potential in biomedical texture analysis where patterns appear at random positions and orientations. LRI operators can be obtained by computing the responses to the discrete rotation of local descriptors, such as Local Binary Patterns (LBP) or the Scale Invariant Feature Transform (SIFT). Other strategies achieve this invariance using Laplacian of Gaussian or steerable wavelets for instance, preventing the introduction of sampling errors during the discretization of the rotations. In this work, we obtain LRI operators via the local projection of the image on the spherical harmonics basis, followed by the computation of the bispectrum, which shares and extends the invariance properties of the spectrum. We investigate the benefits of using the bispectrum over the spectrum in the design of a LRI layer embedded in a shallow Convolutional Neural Network (CNN) for 3D image analysis. The performance of each design is evaluated on two datasets and compared against a standard 3D CNN. The first dataset is made of 3D volumes composed of synthetically generated rotated patterns, while the second contains malignant and benign pulmonary nodules in Computed Tomography (CT) images. The results indicate that bispectrum CNNs allows for a significantly better characterization of 3D textures than both the spectral and standard CNN. In addition, it can efficiently learn with fewer training examples and trainable parameters when compared to a standard convolutional layer.

\end{abstract}


%
\IEEEpeerreviewmaketitle

\vspace{-0.3cm} 

\section{Introduction}

%
%
%
%
\IEEEPARstart{C}{onvolutional} Neural Networks (CNNs) have recently gained a lot of attention as they outperform classical handcrafted methods in almost every computer vision tasks where data scarcity is not an issue. In biomedical image analysis, data are abundant. However, obtaining high quality and consistently labeled images is expensive as data curation and annotation require hours of work from well-trained experts \cite{greenspan2016guest}. Thus, the effective number of training examples is often low. This limitation is usually handled by transfer learning and data augmentation. Transfer learning, the process of fine-tuning a network trained on another task to the task at hand,  is very common for 2D images. For 3D images, however, the lack of very large datasets hinders the availability of pre-trained models. Another approach, data augmentation, refers to the application of geometric transforms and perturbations to the training examples to make the CNN invariant to these distortions~\cite{shorten2019survey}. The cost of data augmentation is a substantial increase in the data size leading to a slower convergence rate and potential waste of trainable parameters. 

A lot of recent research has focused on how to build CNNs that are invariant to these transforms by imposing constraints on the architecture of the network~\cite{CoW2016b, weiler2017learning, andrearczyk2020local, eickenberg2017solid}. The motivation of these approaches is to obviate the need to learn these invariances from the data and their transformation. As a result, an effective reduction of the number of trainable parameters is achieved and, potentially, a reduction of the number of training examples needed for the generalization of the network. 

This work focuses on 3D biomedical texture analysis and on the design of CNNs that are invariant to local 3D rotations, \textit{i.e.}, rotations of individual local patterns. This invariance is obtained using continuously defined Rotation Invariant (RI) descriptors of functions on the sphere. By relying on a continuous-domain formulation, we avoid the difficulties associated with rotations of discretized images~\cite{vivaldi2006arithmetic, ke2014rotation}. Neighborhoods defined by learned radial profiles are used to locally project the image on the solid sphere. These descriptors are used together with a convolution operation to obtain Locally Rotation Invariant~(LRI)\footnote{LRI is used for Locally Rotation Invariant and Local Rotation Invariance interchangeably} operators in the 3D image domain as proposed in \cite{andrearczyk2019exploring}. These types of operators are relevant in biomedical texture analysis where discriminative patterns appear at random positions and orientations. The RI descriptors used in \cite{andrearczyk2019exploring, andrearczyk2019solid, andrearczyk2020local, eickenberg2017solid, weiler20183d} and in the present work are derived from the Spherical Harmonics (SH) decomposition of the kernels. The SHs are the generalization of the Circular Harmonics (CH) to the 2D sphere~\cite{gallier2009}. These two families of functions are intimately linked with Fourier theory, and both decompositions correspond to the Fourier transform of the function defined on the sphere $\mathbb{S}^2$ for the SHs and on the circle $\mathbb{S}^1$ for the CHs. 

To better apprehend the two invariants considered in this work, namely the spectrum and the bispectrum, it is useful to consider them on the circle. The CH expansion of a function $f \in L_2(\mathbb{S}^1)$ for a degree $n$ is computed as $\widehat{f}_n = \frac{1}{2\pi} \int_0^{2\pi}  f(\theta) e^{-\mathrm{j} \theta n} \mathrm{d} \theta$, which is the Fourier series for $2\pi$-periodic functions. For $m,n \in \mathbb{Z}$, the spectrum of the CH expansion is calculated as $s_n(f) = \widehat{f}_n \widehat{f}_n^* = |\widehat{f}_n|^2$ and the bispectrum as $b_{n,m}(f) = \widehat{f}_n \widehat{f}_m \widehat{f}_{n+m}^*$. One readily verifies that for a function $g(\theta) = f(\theta -\theta_0)$ we have for any $m,n \in \mathbb{Z}$ the equalities $s_n(f)=s_n(g)$ and $b_{n,m}(f) = b_{n,m}(g)$, since $\widehat{g}_n = \widehat{f}_n e^{-\mathrm{j}\theta_0 n}$. This means that the spectrum and bispectrum are RI, since a shift $\theta_0$ in the parameter of $f$ is equivalent to a rotation on the circle. The spectrum is the most simple, yet informative, Fourier-based RI quantity. However, it discards the phase between harmonics which contains all the information on how the sinusoids from the expansion add up to form edges and ridges \cite[Chapter 10]{smith1997scientist}. The bispectrum, on the contrary, conserves the phase information~\cite{kakarala2010} and constitutes a more specific pattern descriptor.

The main contributions of this paper are the introduction of a novel image operator based on the Solid Spherical Bispectrum (SSB) that is LRI and a corresponding CNN layer, resulting in a locally rotation invariant CNN. This work builds upon~\cite{andrearczyk2019solid}, where a Solid Spherical Energy (SSE) layer was proposed. 
The radial profiles used to locally project the image on the solid sphere as well as the relative importance of the bispectrum coefficients can be learned end to end with backpropagation.
We experimentally investigate the relevance of the proposed SSB layer for biomedical texture classification. Finally, we study the ability of the SSB-CNN to learn with small amounts of data and compare with a classical CNN.

This manuscript is organized as follows. In Section~\ref{sec:related_work}, we review the main related works. Sections~\ref{sec:notations} to~\ref{sec:SH} describe the nomenclature and the mathematical tools used in this work. The definitions of the spectrum and bispectrum for functions defined on the sphere are reported in Section~\ref{sec:RIsphere} and are drawn from the work of Kakarala and Mao~\cite{kakarala2010}. We recall the theoretical benefits of the bispectrum over the spectrum in Section~\ref{sec:bisp_vs_sp_theory}. In Section~\ref{sec:lri_solid_sphere}, we define the SSE and SSB image operators and state that they are LRI. In Section \ref{sec:implementation}, we discuss the implementation details to integrate these image operators into a convolutional layer, referred to as the SSE or SSB layer.
Sections~\ref{sec:experiments_and_results} and~\ref{sec:discussions} 
detail and discuss the experimental evaluation of the proposed approach.
Conclusions and perspectives are provided in Section~\ref{sec:conclusion}.



\vspace{-0.3cm} 

\section{Related Work}\label{sec:related_work}

\subsection{Rotation Invariant Image Analysis}
Combining LRI and directional sensitivity is not straightforward and is often antagonist in simple designs~\cite{depeursinge2018rotation,andrearczyk2020local}. Several methods exist to combine both properties. Ojala \emph{et al.}~\cite{OPM2002} proposed the Local Binary Patterns (LBP) where they compare values of pixels within a circular neighborhood to the middle pixel. Pixels of the neighborhood are thresholded based on the central pixel to generate a binary code. LRI is achieved by ordering the binary code to obtain the smallest binary number. 

Several LRI filtering approaches were proposed.
Varma and Zisserman~\cite{VaZ2005} used a filter-bank including the same filters at different orientations, where LRI is achieved by max pooling over the orientations. 
Instead of explicitly computing responses (\emph{i.e.} convolving) to oriented filters, steerable filters can be used to improve efficiency~\cite{freeman1991design,Unser2013steerable}. 
The work of Perona~\cite{perona1992steerable} shows the use of steerable filters for LRI edges and junctions analysis.
Dicente \emph{et al.}~\cite{DicenteCid2017} used a filter-bank composed of steerable Riesz wavelets. LRI is obtained by locally aligning the filters to the direction maximizing the gradient of the image. Data-driven steerable filters were used in~\cite{fageot2018principled} as LRI detectors of a given template within an image. Steerable Wavelet Machines (SWMs) were proposed in~\cite{DPW2017}, where task-specific profiles of steerable wavelets are learned using support vector machines. 

Other approaches have been described to obtain invariants without explicitly rotating the descriptors. Such methods relies on moments~\cite{flusser2009moments}
or invariants built from the SH decomposition \cite{kakarala2012bispectrum}. Kakarala and Mao introduced the bispectrum of the SH decomposition in~\cite{kakarala2010} and they demonstrated the superiority of the bispectrum over the spectrum for 3D shape discrimination. In~\cite{kakarala2012bispectrum}, Kakarala showed that the bispectrum has better properties and contains more information than the spectrum, also proving its completeness for functions defined on compact groups. More recently, an extension of the spectral and bispectral invariants was used by Zucchelli \emph{et al.}~\cite{zucchelli2020computational} for the analysis of diffusion Magnetic Resonance Imaging data.

In~\cite{depeursinge2018rotation,eickenberg2017solid}, the authors used the spectrum of the SH expansion to compute LRI operators. Their work shares similarities with the method exposed here. However, our approach is more data-driven since we learn the radial profiles, whereas they rely on handcrafted ones.

\vspace{-0.3cm} 
\subsection{Rotation Equivariance in CNNs}
Recently, several research contributions focused on the explicit encoding of rotation equivariance into CNNs. One group of methods relies on the extension of the classic convolution on the group of translations to groups of symmetries including rotations and reflections. A detailed description of the generalization of the convolution to compact groups is given in~\cite{kondor2018generalization} and to homogeneous spaces in \cite{cohen2019general}. Regarding the application of this generalization, Cohen and Welling~\cite{CoW2016b} used rotations of the filters together with recombinations of the response maps, which is performed according to the rules of group theory and allows equivariance to 2D right-angle rotations.
The same strategy was extended to 3D images in~\cite{winkels2019pulmonary,worrall2018cubenet}.
This 3D group CNN was applied to 3D texture classification in~\cite{andrearczyk2018rotational}. Bekkers \emph{et al.}~\cite{bekkers2018roto} used the convolution on the discretized group of 2D roto-translations. Weiler \emph{et al.}~\cite{weiler2017learning} proposed a CH kernel representation to achieve a more efficient rotation of the filters via steerability, still in the context of the convolution on groups. Cohen and Welling~\cite{cohen2016steerable} used the irreducible representation of the dihedral group to build CNNs that are equivariant to 2D discrete rotations. 

The aforementioned methods offer the possibility to encode the equivariance to virtually any finite group. The 2D rotations group $SO(2)$ can be uniformly discretized by choosing a finite subgroup of $SO(2)$ with an arbitrary large number of elements. This is not anymore the case for 3D rotations since there is only 5 regular convex polyhedrons \cite[Chapter 10]{coxeter1961introduction}.
Therefore, approaches allowing for the propagation of the rotational equivariance without explicitly sampling the different orientations are crucial in 3D. Methods involving CH and SH have been introduced to address this problem. Worrall \emph{et al.}~\cite{WGT2016} used CHs representation of the kernels together with a complex convolution and complex non-linearities to achieve the rotational equivariance. The main drawback is that it generates many channels that must be disentangled to achieve rotation invariance. A SH representation of the kernels was used in~\cite{weiler20183d} to propagate the equivariance as a generalization of~\cite{WGT2016} to 3D images. It is also possible to adapt neural networks to non-Euclidean domains, for instance, to the 2D sphere, where the invariance to rotations plays a crucial role as in~\cite{kondor2018clebsch} and~\cite{cohen2018spherical}. Finally, the group convolution can be extended to more general Lie groups as proposed by Bekkers in~\cite{bekkers2019b}, where CNNs equivariant to roto-translation and scale-translation were implemented.

Most of these methods focused on the propagation of the rotation equivariance throughout the network, whereas we propose lightweight networks discarding this information after each LRI layer, similarly to~\cite{andrearczyk2020local}.

\section{Methods}\label{sec:methods}
\subsection{Notations and Terminology}\label{sec:notations}
We consider 3D images as functions $I \in L_2(\mathbb{R}^3)$, where the value $I(\x) \in \mathbb{R}$ corresponds to the gray level at location $\x = (x_1,x_2,x_3) \in \mathbb{R}^3$. The set of 3D rotation matrices in the Cartesian space is denoted as $SO(3)$. The rotation of an image $I$ is written as  $I(\mathrm{R} \cdot)$, where $\mathrm{R} \in SO(3)$ is the corresponding rotation matrix. 

The sphere is denoted as $\mathbb{S}^2 = \{ \x \in \mathbb{R}^3 : ||\x||_2 = 1\}$. Spherical coordinates are defined as $(\rho,\theta,\phi)$ with radius $\rho \geq 0$, elevation angle $\theta \in [0,\pi]$, and horizontal plane angle $\phi \in [0,2\pi)$. Functions defined on the sphere are written as $f\in L_2(\mathbb{S}^2)$ and are expressed in spherical coordinates. The inner product for $f, g \in L_2(\mathbb{S}^2)$ is defined by $\langle f \,, g \rangle_{L_2(\mathbb{S}^2)} = \int_0^\pi \int_0^{2\pi} f(\theta, \phi) \overline{g(\theta, \phi)} \sin(\theta)\diff\phi \diff\theta$. With a slight abuse of notation, the rotation of a function $f \in L_2(\mathbb{S}^2)$ is written as $f(\mathrm{R}\cdot)$, despite the fact that spherical functions are expressed in spherical coordinates. 

The Kronecker delta $\delta[\cdot]$ is such that $\delta[n]=1$ for $n=0$ and $\delta[n]=0$ otherwise. The Kronecker product is denoted by $\otimes$. The triangle function is referred to as $\text{tri}(x)$ and is defined as $\text{tri}(x)=1-|x|$ if $|x| < 1$ and $\text{tri}(x)=0$ otherwise.
A block diagonal matrix formed by the sub-matrices $\mathrm{A}_i$ is written as $\left[\bigoplus_i \mathrm{A}_i\right]$. The Hermitian transpose is denoted by~$\bm{^\dag}$.

\vspace{-0.3cm} 

\subsection{LRI Operators}\label{sec:lri_def}
This work focuses on image operators $\mathcal{G}$ that are LRI as previously introduced in \cite{andrearczyk2020local}. An operator $\mathcal{G}$ is LRI if it satisfies the three following properties: 
\begin{itemize}
    \item \emph{Locality}: there
    exists $\rho_0 > 0$ such that, for every $\bm{x} \in \R^3$ and every image $I \in L_2(\mathbb{R}^3)$, the
    quantity $\mathcal{G} \{I\} (\bm{x})$ only depends on local
    image values  $I(\bm{y})$ for $\lVert \bm{y} - \bm{x}\rVert
    \leq \rho_0$.
    \item \emph{Global equivariance to translations:} For any $I \in L_2(\mathbb{R}^3)$,
   \begin{equation*}
    \mathcal{G}\{ I (\cdot - \bm{x}_0) \}  = \mathcal{G}\{I\}
    (\cdot - \bm{x}_0) \quad \text{ for any } \bm{x}_0 \in
    \mathbb{R}^3. \label{eq:transinv} \\
   \end{equation*}
    \item  \emph{Global equivariance to rotations:} For any $I \in L_2(\mathbb{R}^3)$,
    \begin{equation*}
       \mathcal{G}\{ I(\mathrm{R}_0 \cdot)  \} =
        \mathcal{G}\{I\} (\mathrm{R}_0\cdot)  \quad \text{ for
       any } \mathrm{R}_0 \in SO(3). \label{eq:rotinv}
   \end{equation*}
\end{itemize}

To reconcile the intuition of LRI with this definition, let us consider a simple scenario where two images $I_1$ and $I_2$ are composed of the same small template $\tau \in L_2(\mathbb{R}^3)$ appearing at random locations and orientations and distant enough to avoid overlaps between them. The locations of the templates $\tau$ are identical for $I_1$ and $I_2$, the difference between the two images being in the local orientation of the templates. These images can be written as
\begin{equation*}
   I_k = \sum_{1 \leq j \leq J} \tau (\mathrm{R}_{j,k} (\cdot - \x_j)),
\end{equation*}
where $J$ is the number of occurrence of the template $\tau$ and $k=1,2$. The local orientation and position of the $j^{\textrm{th}}$ template in image $k$ are represented by $\mathrm{R}_{j,k}$ and $\x_j$, respectively.
If the operator $\mathcal{G}$ is LRI, then   for any $1 \leq j \leq J$ and any rotations $\mathrm{R}_{j,1}$, $\mathrm{R}_{j,2} \in SO(3)$, 
\begin{equation*}
    \mathcal{G}\{I_1\}(\x_j) = \mathcal{G}\{I_{2}\}(\x_j).
\end{equation*}

\begin{figure}
\includegraphics[width=0.48\textwidth]{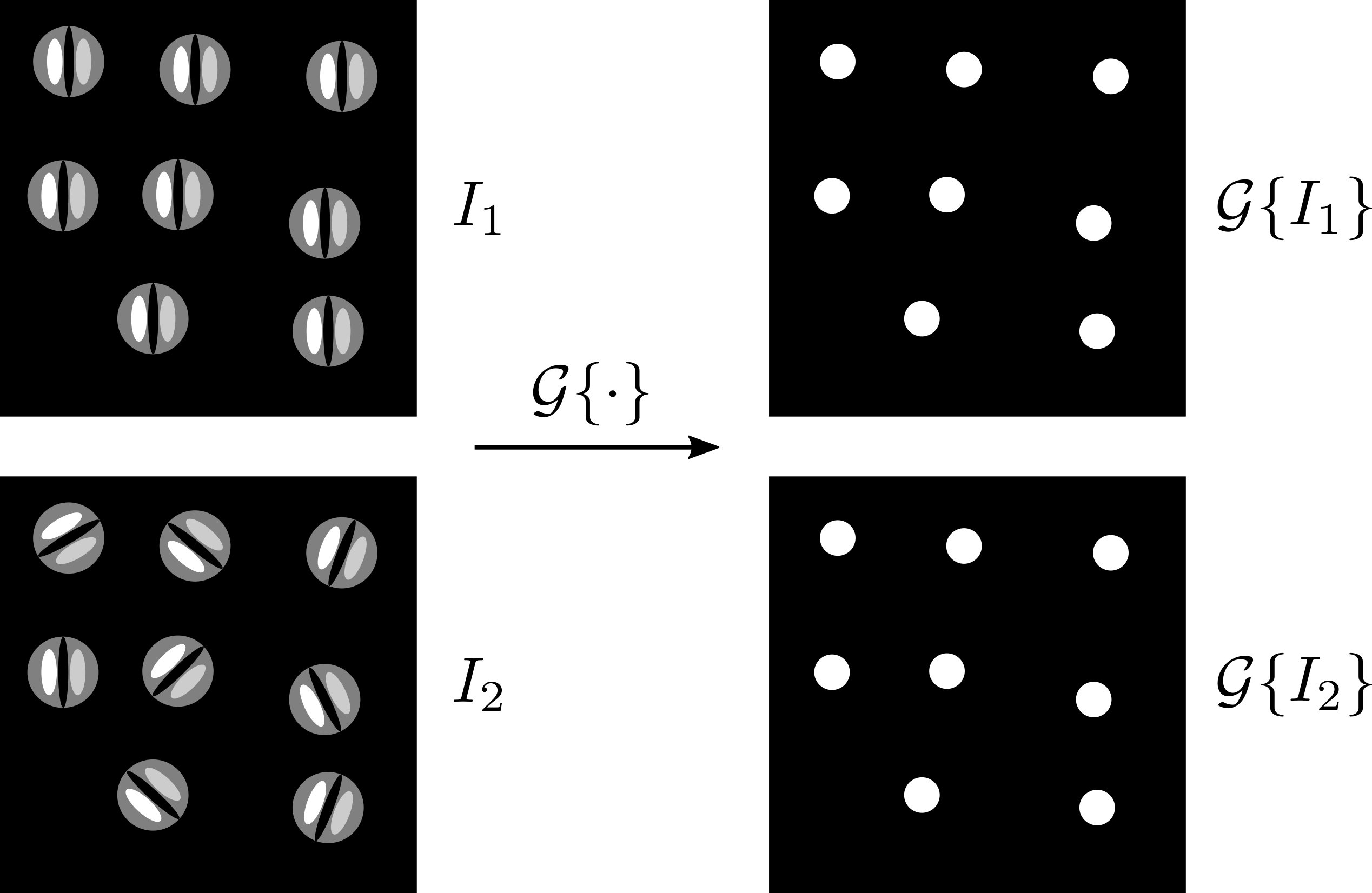}
\caption{Visual representation of the output of a LRI operator. Here, different rotations are applied at the template centers. For the sake of simplicity, only the output values at the template centers are represented.}
\label{fig:lri}
\end{figure}

From the definition of LRI, this equality is required to hold only at the center of the templates. This example is illustrated in Fig. \ref{fig:lri}, where only the responses at the center of the templates are represented.

In this work, the design of LRI operators is obtained in two steps.
First, the image $I \in L_2(\R^3)$ is convolved with SHs modulated by compactly supported radial profiles, referred to as solid SHs. 
The second step involves the computation of RI descriptors for each position.


\vspace{-0.3cm} 

\subsection{Spherical Harmonics}\label{sec:SH}
\begin{figure}
\includegraphics[width=0.49\textwidth]{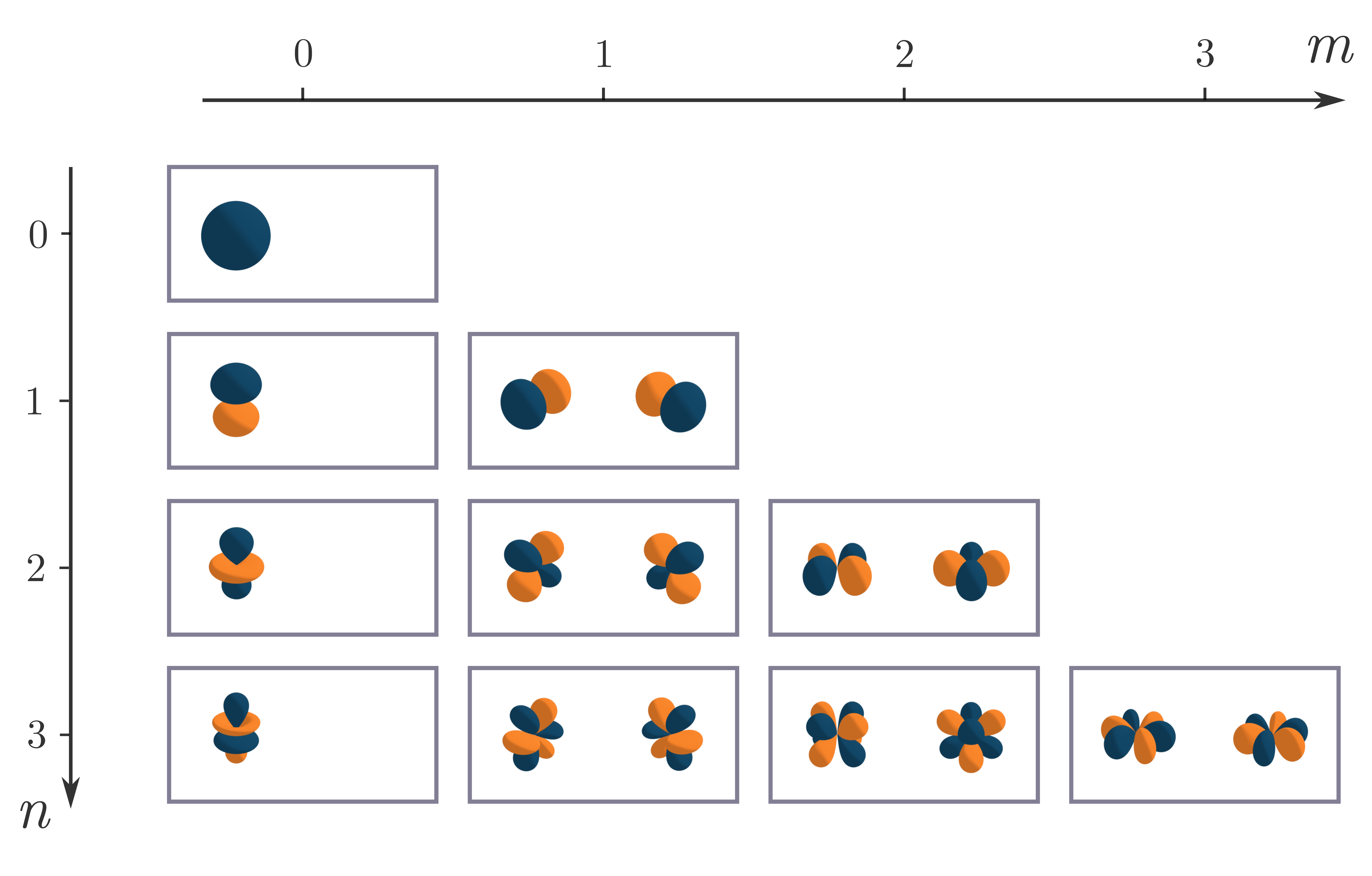}
\caption{Visual representation of the real and imaginary part of the $h(r)Y_n^m(\theta, \phi)$ here $h$ is chosen Gaussian for simplicity. Each box represents a given SH with the real part on the left and the imaginary part on the right. The blue represents positive values and orange negative values. Only the SHs for $m\geq 0$ are represented since we have the following symmetry $Y_n^{-m} = (-1)^m \overline{Y_n^m}$.}
\label{fig:sph_family}
\end{figure}

Any function $f \in L_2(\mathbb{S}^2)$ can be expanded in the form of 
\begin{equation}
\label{eq:sph_exp}
    f(\theta, \phi) = \sum_{n=0}^\infty \sum_{m=-n}^n F_n^m Y_n^m(\theta, \phi),
\end{equation}
where $Y_n^m$ are the so-called SHs for a degree $n \in \mathbb{N}$ and order $m$ with $-n\leq m \leq n$. For their formal definition, see 
\cite[Section 2.5]{depeursinge2018rotation} and for their visual representation, refer to Fig. \ref{fig:sph_family}. The SHs form an orthogonal basis of $L_2(\mathbb{S}^2)$~\cite[Chapter 5.6]{varshalovich1988quantum}. Thus, the expansion coefficients of Eq.~(\ref{eq:sph_exp}) can be computed by projecting $f$ onto
the SH basis using the inner product on the sphere
\begin{equation}
    F_n^m = \langle f\,, Y_n^m \rangle_{L_2(\mathbb{S}^2)}.
\end{equation}
This expansion corresponds to the Fourier transform on the sphere. 
We regroup the coefficients of all orders $m$  for a given degree $n$ as the $1 \times (2n+1)$ vector
\begin{equation}
    \F_n = [F_n^{-n} \ldots F_n^0 \ldots F_n^n],
\end{equation}
called the spherical Fourier vector of degree $n$.
One important property of SHs is their steerability, \textit{i.e.} the rotation of one SH can be
determined by a linear combination of the other SHs of same degree:
\begin{equation} 
\label{eq:steerability_ynm}
     Y_n^m(\mathrm{R}_0\cdot) = \sum_{m'=-n}^n [\mathrm{D}_n(\mathrm{R}_0)]_{m',m} Y_n^{m'},
 \end{equation} 
where $\mathrm{D}_n(\mathrm{R}_0)$ is a unitary matrix known as the Wigner
$\mathrm{D}$-matrix \cite[Chapter 4]{varshalovich1988quantum}.
Therefore, if two functions $f,f' \in L_2(\mathbb{S}^2)$ differ only by a rotation $\mathrm{R}_0 \in SO(3)$, \textit{i.e.} 
$f'=f(\mathrm{R}_0 \cdot)$, their spherical Fourier vectors, $\F_n$ and $\F'_n$,
satisfy the following relation \cite[Section 3, Eq. (5)]{kakarala2010}:
\begin{equation}
\label{eq:wigner_rot}
\F'_n = \F_n \mathrm{D}_n(\mathrm{R}_0).
\end{equation}
This property is similar to the shifting property of the Fourier transform on the real line. In the spherical case, instead of multiplying by a complex exponential, the transform is multiplied by the Wigner $\mathrm{D}$-matrix of degree $n$ associated with the rotation $\mathrm{R}_0$.

\vspace{-0.3cm} 
\subsection{Spherical RI: the Spectrum and the Bispectrum}\label{sec:RIsphere}
With the properties of the spherical Fourier vectors, it is possible to efficiently obtain RI operators for functions defined on the sphere. Two quantities computed from these coefficients will be of interest: the spherical spectrum and the spherical bispectrum. 
\subsubsection{Spectrum} 
The spectrum is a ubiquitous quantity in signal processing and it is well known to provide a source of translation invariant descriptors for periodic functions and functions defined on the real line. In these cases, the spectrum corresponds to the squared modulus of the Fourier transform. Its spherical equivalent, the spherical spectrum, is defined by the averaged squared norm of the spherical Fourier vector $\F_n$:
\begin{equation}
\label{eq:spectrum}
s_n(f) =  \frac{1}{2n+1} \F_n \F_n^{\bm{\dag}} = 
\frac{1}{2n+1}\sum_{m=-n}^n |F_n^m|^2. 
\end{equation}
\subsubsection{Bispectrum}
The bispectrum is defined as in \cite[Section 4 Eq. (24)]{kakarala2010}:
\begin{equation}
\label{eq:bispectrum}
    b^{\ell}_{n,n'}(f) = [\F_n \otimes \F_{n'}] \mathrm{C}_{nn'} \widetilde{\F_\ell}^{\dag},
\end{equation}
where the term $\F_n \otimes \F_{n'}$ is a $1 \times (2n+1)(2n'+1)$ vector, $\mathrm{C}_{nn'}$ is the $(2n+1)(2n'+1) \times (2n+1)(2n'+1)$ Clebsh-Gordan matrix containing the Clebsh-Gordan coefficients, whose definition and main properties are recalled in Appendix~\ref{app:CG}, and $\widetilde{\F_\ell} = [0,  \ldots , 0,  \F_\ell, \, 0, \ldots, 0]$ is a zero-padded vector of size $1 \times (2n+1)(2n'+1)$ containing the spherical Fourier vector of degree $\ell$ with $|n-n'|\leq \ell \leq n+n'$. The zero-padding is performed to match the size of $\mathrm{C}_{nn'}$ and to select only the rows corresponding to the $\ell^{\textrm{th}}$ degree.


The spectrum and the bispectrum are known to be RI. We recall this fundamental result that will be crucial for us thereafter. \\

\vspace{-0.3cm} 
\begin{proposition}\label{prop:ri_spec_bisp_complete}
The spectrum and the bispectrum of spherical functions are RI. This means that, for any rotation $\mathrm{R}_0 \in SO(3)$ and any function $f \in L_2(\mathbb{S}^2)$, we have, for $f' = f( \mathrm{R}_0 \cdot )$,
\begin{equation}
    s_n (f) = s_n (f') \quad \text{and} \quad b^{\ell}_{n,n'} (f) = b^{\ell}_{n,n'} (f') 
\end{equation}
for any $n,n' \geq 0$ and any $|n-n'| \leq \ell \leq n+n'$. \\
\end{proposition}
\vspace{-0.3cm} 

The result that the bispectrum of a spherical function is RI is given in \cite[Theorem 4.1]{kakarala2010}.
Besides, we introduce the following notations:
\begin{equation}
\mathcal{S}\{\F_n\} = s_n(f)
\end{equation} and
\begin{equation}
\mathcal{B}\{\F_n, \F_{n'}, \F_{l}\} = b^\ell_{n, n'}(f).
\end{equation}
These notations highlight that the spectrum coefficient $s_n(f)$ only depends on the $n$th-order Fourier vector $\F_n$, and that the bispectrum coefficient $b_{n,n'}^{\ell}(f)$ only depends on $\F_n$, $\F_{n'}$, and $\F_\ell$.
Moreover, the rotation invariance of the spectrum and bispectrum can be reformulated as
\begin{equation}
\mathcal{S}\{\F_n \mathrm{D}_n(R)\} = \mathcal{S}\{\F_n\}
\end{equation}
and
\begin{equation}
\mathcal{B}\{\F_n \mathrm{D}_n(R) , \F_{n'} \mathrm{D}_{n'}(R), \F_{\ell} \mathrm{D}_\ell(R)\} = \mathcal{B}\{\F_n, \F_{n'}, \F_{\ell}\}.
\end{equation}

\vspace{-0.5cm} 
\subsection{Advantages of the Bispectrum over the Spectrum}\label{sec:bisp_vs_sp_theory}
Despite the simplicity to compute the spherical spectrum, it can be beneficial to use the more complete bispectrum, for which we provide two arguments.
\subsubsection{Inter-Degree Rotations} The spectrum does not take into account the inter-degree rotation. For instance, let us build a function $f'$ from the SH expansion $\F = (\F_0, \F_1, \cdots)$ of the function $f$ as follows: for each degree $n$, we apply a different Wigner $\mathrm{D}$-matrix $\mathrm{D}_n(\mathrm{R}_n)$ with at least one rotation matrix $\mathrm{R}_n$ different from the others. The corresponding SH expansion  $\F'= (\F_0\mathrm{D}_0(\mathrm{R}_0), \F_1\mathrm{D}_1(\mathrm{R_1}), \cdots)$ will have the same spectrum since the Wigner $\mathrm{D}$-matrices are unit matrices (\emph{i.e.}, they do not impact the norm of $\F_n$). 

\subsubsection{Intra-Degree Variations} Another aspect to which the spectrum is insensitive is in the distinction of intra-degree variations. for $n_0\geq1$ fixed, the functions $f=Y_{n_0}^m$ have the same spectrum $s_{n}(f) = \frac{\delta[n-n_0]}{2n_0+1}$ but are not rotation of each other in general (see Fig. \ref{fig:sph_family}).

On the contrary, the bispectrum does not suffer from these limitations (see Section \ref{sec:toyExperiment}). Furthermore, the spectral information is contained in the bispectrum. This can be easily seen as:
\begin{equation}
b^n_{0,n}(f)= \F_0 \F_n \F_n^\dag =F_0^0 s_n(f).
\end{equation}
This illustrates that, given a non-zero mean $\F_0 = F_0^0 \in \R$, we can retrieve the spectral information from the bispectrum. This can appear as a restriction for the bispectrum. However, in practice, it is possible to add a constant to the signal ensuring that $F_0^0$ is non-zero.
The aforementioned properties make the bispectrum a more faithful descriptor and a good substitute of the spectral decomposition.

\vspace{-0.3cm} 
\subsection{LRI on the Solid Sphere $\mathbb{R}^3$}\label{sec:lri_solid_sphere}
The previous sections introduced the theoretical aspects to build RI descriptors for functions defined on the sphere. In this work, we are interested in 3D images, therefore we will use the spherical spectrum and bispectrum in combination with solid SHs. Solid SHs are the multiplication of the SHs by a radial profile to extend them to a 3D volume.
We introduce the following notation for solid SHs evaluated on the Cartesian grid:
\begin{equation}
   \kappa_n^m(\x) = \kappa_n^m (\rho,\theta,\phi) =  h_n(\rho) Y_n^m(\theta, \phi), 
\end{equation}
where $h_{n}$ is a compactly supported radial profile that is shared among the SHs $Y_n^m$ with same degree $n$. In the final network, the radial profiles $h_n$ are learned from the data. 

The image is convolved with the solid SHs and by regrouping the resulting feature maps for each degree, we obtain the spherical Fourier feature maps\footnote{The convolution $(I*\kappa_n^{m})(\x)$  with all the $\kappa_n^m$ is equivalent to a local projection of the image around the position $\x$ to a function defined on the sphere followed by a projection onto the SHs basis. For that reason, we use the same notation as for the spherical Fourier vector of degree $n$. We distinguish the spherical Fourier feature maps by the evaluation over a position $\x$.}:
\begin{equation}
\label{eq:sh_conv}
    \F_n(\x) = [(I*\kappa_n^{m})(\x)]_{m=-n}^{m=n}.
\end{equation}

In other terms, the $m^{\textrm{th}}$ component of $\F_n(\x)$ is $\langle I ( \x - \cdot ) , h_{n} Y_{n}^m \rangle$, and measures the correlation between $I$ centered at $\x$ and the solid SH $\kappa_n^m = h_{n} Y_{n}^m$. 
Thanks to the Fourier feature maps, we introduce the image operators used in this paper in Definition \ref{def:op}.  \\

\begin{definition} \label{def:op}
We define the \emph{SSE image operator} $\mathcal{G}^\text{SSE}_n$ of degree $n \geq 0$ as
\begin{equation}
\label{eq:gsse}
    \mathcal{G}^\text{SSE}_n\{I\}(\x) = \mathcal{S}\{ \F_n (\x)\}
\end{equation}
for any $I \in L_2(\R^3)$ and $\x \in \R^3$.
 Similarly, we define the \emph{SSB image operator} $\mathcal{G}^\text{SSB}_{n,n',\ell}$ associated with degrees $n,n' \geq 0$ and $|n-n'| \leq \ell \leq n+n'$ as 
\begin{equation}
\label{eq:gssb}
    \mathcal{G}^\text{SSB}_{n,n',\ell}\{I\}(\x) = 
    \mathcal{B}\{ \F_n(\x), \F_{n'}(\x), \F_{\ell}(\x) \},
\end{equation}
for any $I \in L_2(\R^3)$ and $\x \in \R^3$. \\
\end{definition}

The SSE image operators have been considered in \cite{andrearczyk2020local}, where it was proven to be LRI in Appendix D. We recall this result and extend it to SSB image operators in the following proposition, whose proof is given in Appendix \ref{app:LRIproof}. \\

\vspace{-0.3cm} 

\begin{proposition} \label{prop:LRI}
The SSE and SSB image operators are globally equivariant to translations and rotations. When the radial profiles $h_n$ are all compactly supported, these operators are therefore LRI in the sense of Section \ref{sec:lri_def}.
\end{proposition}
\vspace{-0.3cm}

\subsection{Implementation of the LRI layers}\label{sec:implementation}
In this section, we report the implementation details of our LRI design.

\subsubsection{Parameterization of the Radial Profiles}

The radial profiles are parameterized as a linear combination of radial functions 
\begin{equation} \label{eq:hqn}
    h_{q,n}(\rho) = \sum_{j=0}^{J} w_{q,n,j} \psi_j(\rho),
\end{equation} 
where the  $w_{q,n,j}$ are the trainable parameters of the model.
In \eqref{eq:hqn}, $h_{q,n}$ is the $q^\textrm{th}$ radial profile associated to the degree $n$. The index $q$ controls the number of output streams in the layer.
The index $j=0,\ldots,J$ controls the radial components of the filter. The radial functions are chosen as $\psi_j(\rho)=\text{tri}(\rho-j)$.

\subsubsection{Number of Feature Maps}
The image is convolved with the kernels $\kappa_{q,n}^m$ to obtain the spherical Fourier feature maps $\{\F_{q,n}(\x)\}_{n=0,\ldots,N}^{q=1,\ldots,Q}$. Here, $Q$ is the number of output streams of the layer and $N$ is the maximal degree of the SH decomposition. These feature maps are combined according to Eq. (\ref{eq:gsse}) and (\ref{eq:gssb}) resulting in $\mathcal{G}^\text{SSE}_{q,n}\{I\}(\x)$ or $\mathcal{G}^\text{SSB}_{q,n,n',l}\{I\}(\x)$ respectively.
In the following, we discuss the number of feature maps generated for only one output stream, thus we drop the index $q$.

In the case of the operator $\mathcal{G}^\text{SSE}_n$, the number of generated feature maps is $N+1$. For the $\mathcal{G}^\text{SSB}_{n,n',\ell}$ operator, the total number of features maps is $\mathcal{O}(N^3)$. It is actually not necessary to compute all the bispectrum coefficients, some of them being redundant due to the following properties. First, for each $n$, $n'$ and $\ell$, the bispectral components $b_{n,n'}^\ell(f)$ and $b_{n',n}^\ell(f)$ are proportional independently of $f$~\cite[Theorem 4.1]{kakarala2010}. Hence, we choose to compute the components only for $n\leq n'$ and $0\leq n+n'\leq N$. Second, even though the bispectrum is complex-valued, when $f$ is real, $b^\ell_{n,n'}(f)$ is either purely real or purely imaginary if $n+n'+\ell$ is even or odd respectively~\cite[Theorem 2.2]{kakarala2011viewpoint}. Thus, we can map it to a real-valued scalar. In our design, we take either the real or the imaginary part depending on the value of the indices $n,n',\ell$.

Even with these two properties the number of feature maps for the $\mathcal{G}^\text{SSB}_{n,n',\ell}$ operator still follows a polynomial of degree 3 (see Table~\ref{tab:comp_num_feature_map} for the first values), but for low $N$ it still reduces greatly this number. Moreover, the maximal degree $N$ for the SH expansion cannot be taken arbitrarily large as the kernels are discretized~\cite{andrearczyk2020local}.  The upper bound for $N$ is given by $N \leq \frac{\pi c}{4}$, where $c$ is the diameter of the kernel. This condition can be regarded as the Nyquist frequency for the SH expansion. As an example, $N=7$ is the maximal value for a kernel of size $9\times9\times9$.

\begin{table}[h]
\caption{Number of feature maps obtained for the $\mathcal{G}^\text{SSE}_n$ and $\mathcal{G}^\text{SSB}_{n,n',\ell}$ operators in function of the maximal degree $N$.}
\label{tab:comp_num_feature_map}
\centering
\begin{tabular}{@{}lrrrrrrrr@{}} \toprule
$N$  & 0 & 1 & 2 & 4 & 6  & 8  & 10 & 100   \\ \midrule
Spectrum   & 1 & 2 & 3 & 5 & 7 & 9  & 11 & 101   \\
Bispectrum & 1 & 2 & 5 & 14 & 30& 55 & 91 & 48127 \\ \bottomrule
\end{tabular}
\end{table}

\subsubsection{Discretization}
The kernels $\kappa_{q,n}^m = h_{q,n} Y_n^m$ are discretized by evaluating them on a Cartesian grid. For more details about the discretization, see \cite[Section 2.4]{andrearczyk2019exploring}.

\vspace{-0.3cm} 

\section{Experiments and Results}\label{sec:experiments_and_results}
Section~\ref{sec:toyExperiment} illustrates the differences between the spherical spectrum and bispectrum with two toy experiments designed to fool the spectrum. Then, we compare the classification performance of three different CNNs (SSE, SSB and standard) detailed in Section~\ref{sec:network_architecture} on the two datasets described in Section~\ref{sec:datasets} in terms of accuracy (Section~\ref{sec:exp_perf}) and generalization power (Section~\ref{sec:learningCurves}) \emph{i.e.} the number of training samples required to reach the final accuracy.

\vspace{-0.3cm} 
\subsection{Comparing Local Spectrum to Bispectrum Representations}\label{sec:toyExperiment}
Two toy experiments are conducted to highlight the differences in terms of the representation power of the spherical spectrum and bispectrum. The first experiment is designed to show that the spectrum is unable to discriminate between patterns that only differ in terms of rotations between degrees. The second experiment illustrates that the spectrum cannot capture differences within the same degree. These two experiments are done in the 3D image domain to show the applicability of the spherical spectrum and bispectrum of the solid SHs and to be as close as possible to the final application.

\begin{figure}[h]
\subfloat[Spectrum]{
\begin{tikzpicture}
\begin{axis}[
    ybar,
    enlargelimits=0.1,
    legend style={at={(0.5\linewidth,-0.2)},
      anchor=north,legend columns=-1, font=\tiny},
    symbolic x coords={$s_0$,$s_1$,$s_2$,$s_3$, $s_4$},
    xtick=data,
    height=5cm,
    width=5cm,
    bar width=0.2cm,
    font=\tiny,
    ]
\addplot+[error bars/.cd, y dir=both,y explicit] coordinates
{($s_0$,0.00157164443920830)+-(0.00200772953527793,0.00200772953527793)
($s_1$,0.995507766356341)+-(0.0444060471242965,0.0444060471242965)
($s_2$,0.990281913008063)+-(0.0279315860516880,0.0279315860516880)
($s_3$,0.998082259068930)+-(0.0310790272735265,0.0310790272735265)
($s_4$,0.00141246579581384)+-(0.0006,0.0006)};
\addplot+[color=myorange, fill=mylightorange, error bars/.cd, y dir=both,y explicit] coordinates
{($s_0$,0.00179089174427811)+-(0.00205355042270627,0.00205355042270627) 
($s_1$,1.00477795233041)+-(0.0510030262118858,0.0510030262118858)
($s_2$,0.999704045285163)+-(0.0385291015950738,0.0385291015950738)
($s_3$,0.999497332229525)+-(0.0272879097571221,0.0272879097571221)
($s_4$,0.00172596193723601)+-(0.000884775393148029,0.000884775393148029)};
\end{axis}
\end{tikzpicture}
} 
\subfloat[Bispectrum]{
\begin{tikzpicture}
\begin{axis}[
    at={(0.45\linewidth, 0)},
    ybar,
    enlargelimits=0.15,
    legend style={at={(0.5,-0.15)},
      anchor=north,legend columns=-1, font=\tiny},
    symbolic x coords={$b^1_{1,2}$,$b^2_{1,2}$,$b^3_{1,2}$,
    $b^2_{1,3}$},
    xtick=data,
    height=5cm,
    width=5cm,
    bar width=0.2cm,
    font=\tiny,
    ]
\addplot+[error bars/.cd, y dir=both,y explicit] coordinates
{($b^1_{1,2}$,1.54986895219390)+-(0.153894362386040,0.153894362386040) 
($b^2_{1,2}$,0)+-(0.,0.)
($b^3_{1,2}$,6.26968554144516)+-(0.240570081208577,0.240570081208577)
($b^2_{1,3}$,5.29885141116224)+-(0.203318827693813,0.203318827693813)};
\addplot+[color=myorange, fill=mylightorange, error bars/.cd, y dir=both,y explicit] coordinates {($b^1_{1,2}$,1.23363451489581)+-(0.179168140493069,0.179168140493069) 
($b^2_{1,2}$,0)+-(0.0,0.0)
($b^3_{1,2}$,-4.91308009918514)+-(0.203532701821435,0.203532701821435)
($b^2_{1,3}$,-4.15231054964832)+-(0.172016528920777,0.172016528920777)};

\end{axis}
\end{tikzpicture}
} 
\caption{Experiment with distinct inter-degree rotations. Spherical spectral (left) and bispectral (right) decomposition of the two classes. The blue bars represent the decomposition for the first class $f$ 
and the orange bars for the second class $f'$ ($\F'_i=\F_i \D_i(R_i)$, $i=1,2,3$). Note that only a subset of the bispectral components is displayed. It can be observed that the spectrum cannot distinguish between $f$ and $f'$, and that the bispectrum can.
}
\label{fig:results_scalar1}
\end{figure}
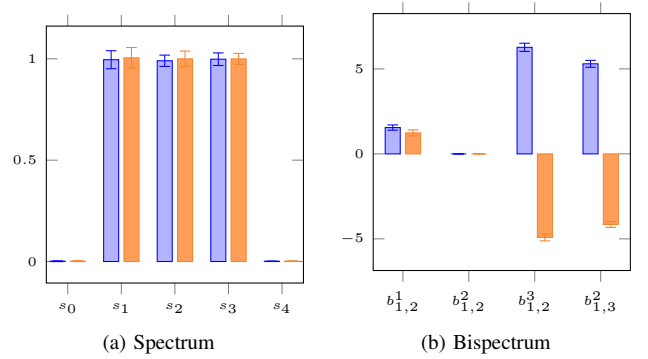

The images are obtained by evaluating $h(\rho) \sum_{n=0}^N \sum_{m=-n}^{m=n}F_n^m Y_m^n(\theta, \phi)$ on a 3D Cartesian grid of $32\times32\times32$ with $h$ defined as an isotropic Simoncelli wavelet profile \cite{portilla2000parametric}. This first experiment investigates the capability of the spectrum and bispectrum to discriminate between functions with distinct inter-degree rotations. Representatives $f$ and $f'$ of the two classes are obtained by summing the SHs described by their repsective spherical Fourier transform $\F$ and $\F'$. $\F$ is composed of $\F_1 = [1 , \mathrm{j},1]$, $\F_2 = [1, -1 ,1,1,1]$, $\F_3 = [1 ,-1 ,1, \mathrm{j},1, 1, 1]$ and $\F_n = \boldsymbol{0}$ for any $n\neq 1,2,3$. The coefficients are chosen to ensure that the images are real and that the spherical spectrum $s_n(f)=1$ for $n=1,2,3$. The spherical decomposition $\F'$ of the second class is computed as $\F'_1=\F_1 \D_1(\mathrm{R}_1)$, $\F'_2=\F_2 \D_2(\mathrm{R}_2)$ and  $\F'_3=\F_3 \D_3(\mathrm{R}_3)$, where $\mathrm{R}_1$, $\mathrm{R}_2$ and $\mathrm{R}_3$ are distinct 3D rotations. This allows to combine the different degrees with different rotations resulting in a function $f'$ with spectrum $s_{n}(f') = s_n(f)$ for all $n$ but $f'\neq f$. Moreover, for each class, 50 distinct instances are created by adding Gaussian noise and randomly rotating the images. The random rotations are drawn from a uniform distribution over the 3D rotations and then we use the associated Wigner-$\mathrm{D}$ matrices to rotate the instances. This time, the same rotation is applied to all degrees. The bispectrum and spectrum are calculated using only the responses to the spherical filters at the origin voxel of the images and the results are presented in Fig. \ref{fig:results_scalar1}. Note that only a subset of distintive coefficients of the bispectrum is shown. The results indicate that the spectrum cannot detect inter-degree rotations, whereas the bispectrum can.

\begin{figure}[h]
\subfloat[Spectrum]{
\begin{tikzpicture}
\begin{axis}[
    ybar,
    enlargelimits=0.1,
    symbolic x coords={$s_0$,$s_1$,$s_2$,$s_3$, $s_4$},
    xtick=data,
    height=5cm,
    width=5cm,
    bar width=0.2cm,
    font=\tiny,
    ]
\addplot+[error bars/.cd, y dir=both,y explicit] coordinates
{($s_0$,0.00157161884978187)+-(0.00200679250614507,0.00200679250614507)
($s_1$,0.994206747907757)+-(0.043,0.043)
($s_2$,0.999181996439430)+-(0.031,0.031)
($s_3$,1.00433690101238)+-(0.030,0.030)
($s_4$,0.00141306630623314)+-(0.0006,0.0006)};
\addplot+[color=myorange, fill=mylightorange, error bars/.cd, y dir=both,y explicit] coordinates
{($s_0$,0.00179108623023528)+-(0.0021,0.0021) 
($s_1$,0.990150993901337)+-(0.053,0.053)
($s_2$,0.997505506247808)+-(0.030,0.030)
($s_3$,1.00258395242995)+-(0.034,0.034)
($s_4$,0.00172138107915893)+-(0.0009,0.0009)};
\end{axis}
\end{tikzpicture}
}
\subfloat[Bispectrum]{
\begin{tikzpicture}
\begin{axis}[
    ybar,
    enlargelimits=0.15,
    symbolic x coords={$b^2_{1,1}$,$b^1_{1,2}$,$b^2_{1,2}$,
    $b^3_{1,2}$},
    xtick=data,
    height=5cm,
    width=5cm,
    bar width=0.2cm,
    font=\tiny,
    ]
\addplot+[error bars/.cd, y dir=both,y explicit] coordinates
{($b^2_{1,1}$,-5.43)+-(0.24,0.24) 
($b^1_{1,2}$,-4.21)+-(0.18,0.18)
($b^2_{1,2}$,0)+-(0,0)
($b^3_{1,2}$,7.91)+-(0.23,0.23)};
\addplot+[color=myorange, fill=mylightorange, error bars/.cd, y dir=both,y explicit] coordinates
{($b^2_{1,1}$,4.67)+-(0.26,0.26) 
($b^1_{1,2}$,3.62)+-(0.2,0.2)
($b^2_{1,2}$,0)+-(0,0)
($b^3_{1,2}$,7.1916)+-(0.24,0.24)};
\end{axis}
\end{tikzpicture}
} 
\caption{Experiment with intra-degree variations. Spherical spectral (left) and bispectral (right) decomposition of the two classes. The blue bars represent the decomposition for the first class $f$ 
and the orange bars for the second class $f'$.
Note that only a subset of the bispectral components is displayed.  It can be observed that the spectrum cannot distinguish between $f$ and $f'$, and that the bispectrum can.}
\label{fig:results_scalar2}
\end{figure}
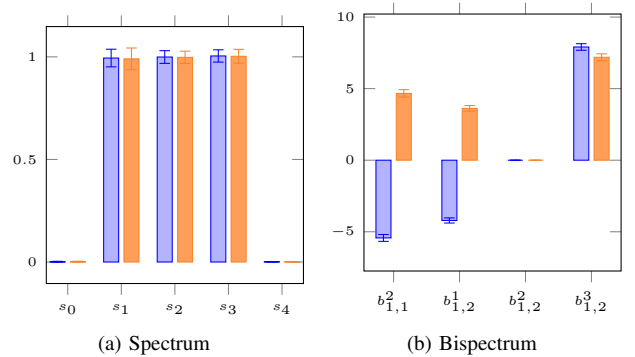

In the next experiment, instead of applying a distinct rotation to each degree, we choose orders that are different within the same degree. For the first class $f$, we use only the order $m=0$ and for the second class $f'$ the orders $m=n,-n$. This choice is motivated by their differences in shape as represented in Fig. \ref{fig:sph_family}. The spherical Fourier transform $\F$ of the first class is chosen to be $\F_1 = [0 ,\sqrt{3} \mathrm{j} ,0]$, $\F_2 = [0 ,0,\sqrt{5},0,0]$, $\F_3 = [0,0,0,\sqrt{7} \mathrm{j},0,0,0]$ and $\F_n = 0$ for any $n\neq 1,2,3$. The spherical decomposition $\F'$ of the second class is $\F'_1 = [\sqrt{3/2},0,\sqrt{3/2}]$, $\F'_2 = [ \sqrt{5/2},0,0,0,\sqrt{5/2}]$, $\F'_3 = [\sqrt{7/2},0,0,0,0,0,\sqrt{7/2}]$. The coefficients are chosen to obtain a spherical spectrum of 1 for $n=1$ , $n=2$ and $n=3$ and to generate real images. The results in Fig. \ref{fig:results_scalar2} show that the bispectrum can discriminate between the two classes even though they have the same spectrum.

\vspace{-0.3cm} 
\subsection{Datasets}\label{sec:datasets}
To evaluate the performance of the proposed LRI operators, we use both a synthetic and a medical dataset. The synthetic dataset constitutes a controlled experimental setup and contains two classes with 500 volumes each of size $32\times32\times32$ for each class. They are generated by placing two types of patterns with a size of $7\times7\times7$, namely a binary segment and a 2D cross with the same norm, at random 3D orientations and random locations with possible overlap. The number of patterns per volume is randomly set to $\lfloor d \cdot(\frac{s_v}{s_p})^3\rfloor$, where $s_v$ and $s_p$ are the sizes of the volume and of the pattern, respectively and the density $d$ is drawn from a uniform distribution in the range $[0.1,0.5]$. The two classes vary by the proportion of the patterns, \textit{i.e.} 30\% segments with 70\% crosses for the first class and vice versa for the second class. 800 volumes are used for training and the remaining 200 for testing. Despite the simplicity of this dataset, some variability is introduced by the overlapping patterns and the linear interpolation of the 3D rotations.

\begin{figure}[h]
\subfloat[Benign]{
\includegraphics[width=0.24\textwidth]{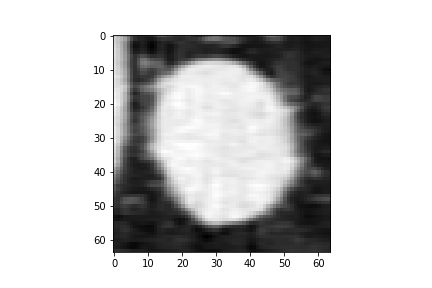}
}
\subfloat[Malignant]{
\includegraphics[width=0.24\textwidth]{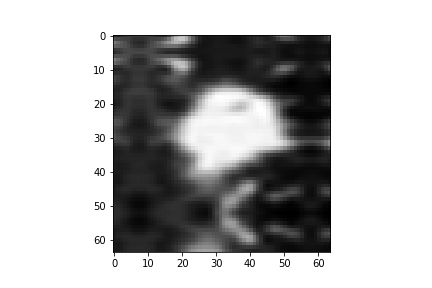}
} 
\caption{Slices drawn from the NLST dataset showing a benign pulmonary nodule
and a malignant one.}
\label{fig:ex_nlst}
\end{figure}

The second dataset is a subset of the American National Lung Screening Trial (NLST) that was annotated by radiologists at the University Hospitals of Geneva (HUG)~\cite{andrearczyk2020local}. The dataset includes 485 pulmonary nodules from distinct patients in CT, among which 244 benign and 241 malignant. We pad or crop the input volumes (originally with sizes ranging from $16\times16\times16$ to $128\times128\times128$) to the size $64\times64\times64$. We use balanced training and test splits with 392 and 93 volumes respectively. Examples of 2D slices of the lung nodules are illustrated in Fig. \ref{fig:ex_nlst}. The Hounsfield units are clipped in the range $[-1000,400]$, then normalized with zero mean and unit variance (using the training mean and variance).

\vspace{-0.3cm} 

\subsection{Network Architecture}\label{sec:network_architecture}
This work uses the network architecture proposed in \cite{andrearczyk2019exploring}. The first layer consists of the LRI layer implemented as described in Section \ref{sec:implementation}. The obtained responses are aggregated using spatial global average pooling, similarly to \cite{AnW2016}. This pooling aggregates the LRI operator responses into a single scalar per feature map and is followed by one Fully Connected (FC) layer.  For the nodule classification experiment, we average the responses inside the nodule masks instead of across the entire feature maps to remove the effect of the size allowing the network to focus on the textural content of the nodule. The final softmax FC layer is connected directly with a cross-entropy loss. Standard Rectified Linear Units (ReLU) activations are used. The two different types of LRI networks are referred to as SSE-CNN and SSB-CNN when the LRI layer uses the $\mathcal{G}^\text{SSE}$ or $\mathcal{G}^\text{SSB}$ operator respectively.

The networks are trained using an Adam optimizer with $\beta_1=0.99$ and $\beta_2=0.9999$ and a batch size of 8. Other task-specific parameters are: for the synthetic experiment (kernel size $7\times7\times7$, stride 1, 2 filters and 50,000 iterations), for the nodule classification experiment (kernel size $9\times9\times9$, stride 2, 4 filters and 10,000 iterations).
The initial values of the trainable weights in \eqref{eq:hqn} are drawn independently from a Gaussian distribution as $w_{q,n,j} \sim \mathcal{N}(0,\,1)$ and the biases are initialized to zero. This initialization is inspired by \cite{he2015delving,weiler2017learning} in order to avoid vanishing and exploding activations and gradients.

We compare the proposed CNNs to a network with the same architecture but with a standard 3D convolutional layer and varying numbers of filters, referred to as Z3-CNN.

\vspace{-0.3cm} 
\subsection{Classification Performance of the SSB-, SSE- and Z3-CNN}\label{sec:exp_perf}
Here, we evaluate the classification performance of both the SSE-CNN and SSB-CNN on the two datasets described in Section \ref{sec:datasets}. The accuracies of both designs are computed with 10 different initializations for varying maximal degrees $N$.
Confidence Intervals (CI) at $95\%$ and mean accuracies are reported in Fig. \ref{fig:res_perf_synth} and \ref{fig:res_perf_NLST} for the synthetic and NLST datasets respectively. On both datasets, the SSB-CNN outperforms the two other networks. 
To exclude the possibility that this performance gain is simply due to a higher number of feature maps, we trained a SSE-CNN on the synthetic dataset with maximal degree $N=2$ and 4 kernels in the first layer instead of 2. This amounts to a total of 12 feature maps after the first layer.
This model achieves $0.9075 \pm 0.006$ of accuracy and is still significantly outperformed by the SSB-CNN with maximal degree $2$ and 2 kernels, which has 10 feature maps after the first layer and obtains an accuracy of $0.924 \pm 0.008$ (Fig.~\ref{fig:res_perf_synth}). 
One important remark is that both LRI networks contain fewer parameters than the Z3-CNN. For instance in the NLST experiment, the SSB- and SSE-CNN have 330 and 222 parameters respectively for a maximal degree $N=4$ against 7322 parameters for the Z3-CNN.



\begin{figure}
\begin{tikzpicture}
\begin{axis}[legend pos=south east, ymin=0.7, ymax=0.95, ylabel=Accuracy] 

\addplot [color=blue,only marks, mark=*]
 plot [error bars/.cd, y dir = both, y explicit]
 table[row sep=crcr, x index=0, y index=1, y error index=2,]{
 0 0.7355 0.1100561353 \\
 1 0.881 0.02800165191\\
 2 0.924 0.007689855677 \\
 3 0.923 0.003844927839 \\
 4 0.9155 0.006183689349 \\
};

\addplot [color=myorange,only marks, mark=*]
 plot [error bars/.cd, y dir = both, y explicit]
 table[row sep=crcr, x index=0, y index=1, y error index=2,]{
 0 0.738 0.07486035234\\
 1 0.8575 0.03151045897\\
 2 0.9 0.008761297018\\
 3 0.9085 0.007163497682\\
 4 0.9025 0.00590139339\\
};

\addplot [dashed, no markers, thick, mygreen]
 plot [error bars/.cd, y dir = both, y explicit]
 table[row sep=crcr, x index=0, y index=1, y error index=2,]{
 0 0.875 0.0\\
 4 0.875 0.0\\
};
\addplot [dashed, no markers, mygreen]
 plot [error bars/.cd, y dir = both, y explicit]
 table[row sep=crcr, x index=0, y index=1, y error index=2,]{
 0 0.865 0.0\\
 4 0.865 0.0\\
};
\addplot [dashed, no markers, mygreen]
 plot [error bars/.cd, y dir = both, y explicit]
 table[row sep=crcr, x index=0, y index=1, y error index=2,]{
 0 0.885 0.0\\
 4 0.885 0.0\\
};
\addlegendentry{SSB-CNN}
\addlegendentry{SSE-CNN}
\addlegendentry{Z3-CNN}

\end{axis}
\node[] at (0.55,-0.7) {{\color{blue}$22$}/{\color{myorange}$22$}};
\node[] at (1.9,-0.7) {{\color{blue}$42$}/{\color{myorange}$42$}};
\node[] at (3.3,-0.7) {{\color{blue}$74$}/{\color{myorange}$62$}};
\node[] at (4.55,-0.7) {{\color{blue}$112$}/{\color{myorange}$82$}};
\node[] at (6.05,-0.7) {{\color{blue}$156$}/{\color{myorange}$102$}};
\node[] at (-1,-0.23) {$N$};
\node[] at (-1,-0.7) {$\#$ param.};

\end{tikzpicture}
\caption{Classification accuracies and numbers of parameters on the synthetic dataset for varying maximal degrees $N$. The error bars represent the CIs at $95\%$. The accuracy of the Z3-CNN with 10 filters is $0.875 \pm 0.011$ with 3462 trainable parameters and is represented by the green dashed lines.}
\label{fig:res_perf_synth}
\end{figure}
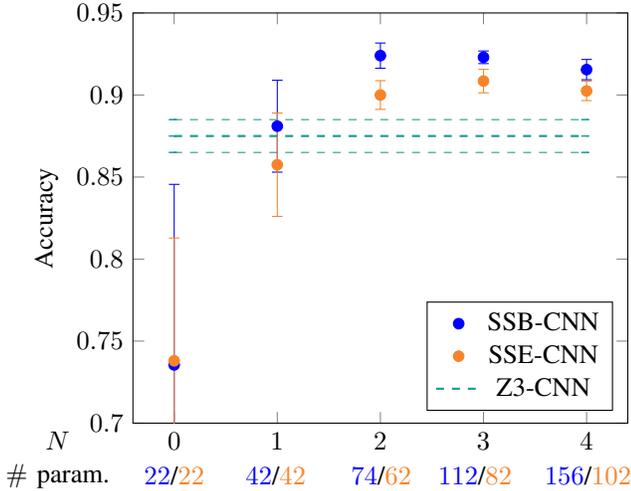
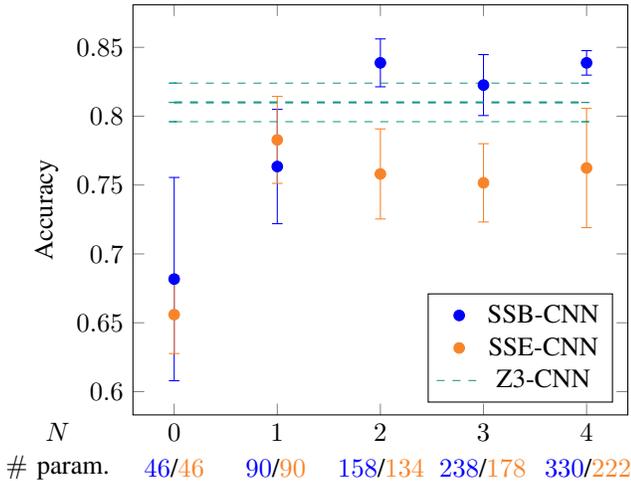
\begin{figure}
\begin{tikzpicture}
\begin{axis}[legend pos=south east, ylabel=Accuracy]

\addplot [color=blue,only marks, mark=*,domain=0.7:1]
 plot [error bars/.cd, y dir = both, y explicit]
 table[row sep=crcr, x index=0, y index=1, y error index=2,]{
0	0.6817204305 0.07379698421\\
1	0.7634408603 0.0415020181\\
2	0.8387096775 0.01738991601\\
3	0.8225806452 0.0221307793\\
4	0.8387096774 0.00888196779\\
};

\addplot [color=myorange,only marks, mark=*,domain=0.7:1]
 plot [error bars/.cd, y dir = both, y explicit]
 table[row sep=crcr, x index=0, y index=1, y error index=2,]{
0	0.6559139789 0.02826224835 \\
1	0.782795699 0.03156953398 \\
2	0.7580645163 0.03268475632 \\
3	0.7516129034 0.02842460968 \\
4	0.7623655915 0.04327776198 \\
};

\addplot [dashed, no markers, mygreen]
 plot [error bars/.cd, y dir = both, y explicit]
 table[row sep=crcr, x index=0, y index=1, y error index=2,]{
 0 0.824 0.0\\
 4 0.824 0.0\\
};
\addplot [dashed, no markers, thick, mygreen]
 plot [error bars/.cd, y dir = both, y explicit]
 table[row sep=crcr, x index=0, y index=1, y error index=2,]{
 0 0.810 0.0\\
 4 0.810 0.0\\
};
\addplot [dashed, no markers, mygreen]
 plot [error bars/.cd, y dir = both, y explicit]
 table[row sep=crcr, x index=0, y index=1, y error index=2,]{
 0 0.796 0.0\\
 4 0.796 0.0\\
};
\addlegendentry{SSB-CNN}
\addlegendentry{SSE-CNN}
\addlegendentry{Z3-CNN}

\end{axis}
\node[] at (0.55,-0.7) {{\color{blue}$46$}/{\color{myorange}$46$}};
\node[] at (1.9,-0.7) {{\color{blue}$90$}/{\color{myorange}$90$}};
\node[] at (3.3,-0.7) {{\color{blue}$158$}/{\color{myorange}$134$}};
\node[] at (4.65,-0.7) {{\color{blue}$238$}/{\color{myorange}$178$}};
\node[] at (6.05,-0.7) {{\color{blue}$330$}/{\color{myorange}$222$}};
\node[] at (-1,-0.23) {$N$};
\node[] at (-1,-0.7) {$\#$ param.};

\end{tikzpicture}%
\caption{Classification accuracies and numbers of parameters on the NLST dataset for varying maximal degrees $N$.
The error bars represent the CIs at $95\%$.
The accuracy of the Z3-CNN with 10 filters is $0.810 \pm 0.014$
with 7322 trainable parameters and is represented by the green dashed lines.}
\label{fig:res_perf_NLST}
\end{figure}

\vspace{-0.3cm} 

\subsection{Learning Curves of the SSB-, SSE- and Z3-CNN}\label{sec:learningCurves}

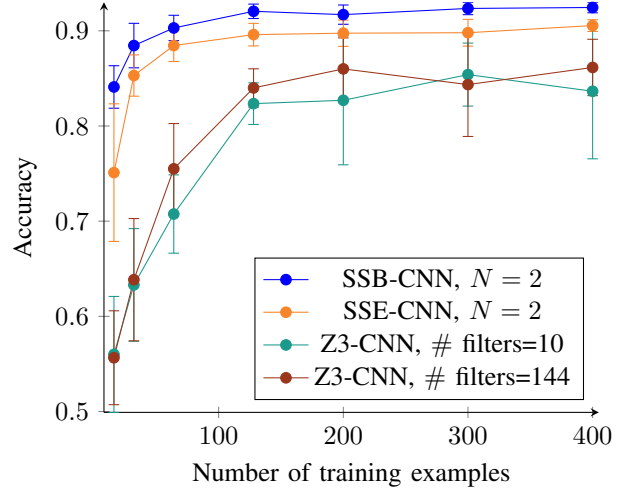
\begin{figure}
\begin{tikzpicture}
\begin{axis}[
    axis lines = left,
    legend pos=south east,
    xmin=8, xmax=405,
    ylabel=Accuracy,
    xlabel= Number of training examples
]

\addplot [color=blue, mark=*,domain=0.7:1]
 plot [error bars/.cd, y dir = both, y explicit]
 table[row sep=crcr, x index=0, y index=1, y error index=2,]{
16   0.841   0.02232227531  \\
32   0.8845  0.02337461542  \\
64   0.903   0.01333835333  \\
128  0.9205  0.007568334006 \\
200  0.917   0.01011998506  \\
300  0.9235  0.006544972563 \\
400  0.9245  0.005569529117 \\
};
\addplot [color=myorange, mark=*,domain=0.7:1]
 plot [error bars/.cd, y dir = both, y explicit]
 table[row sep=crcr, x index=0, y index=1, y error index=2,]{
16 0.751  0.07228477352  \\
32 0.853  0.02168073911  \\
64 0.8845  0.01674783247  \\
128 0.896   0.0118528956   \\
200 0.8975  0.01382771881  \\
300 0.898   0.01405726462  \\
400 0.9055 0.006130632989 \\
};

\addplot [color=mygreen, mark=*,domain=0.7:1]
 plot [error bars/.cd, y dir = both, y explicit]
 table[row sep=crcr, x index=0, y index=1, y error index=2,]{
16	0.56 0.06090343617 \\
32	0.633 0.05918688487 \\
64	0.7075 0.04112553633 \\
128	0.8235 0.02189542377 \\
200	0.827 0.06779499158 \\
300	0.854 0.03310135262 \\
400	0.8365 0.07098455065 \\
};

\addplot [color=myred, mark=*,domain=0.7:1]
 plot [error bars/.cd, y dir = both, y explicit]
 table[row sep=crcr, x index=0, y index=1, y error index=2,]{
16	0.5565 0.04932037049 \\
32	0.6385 0.06423605343 \\
64	0.755 0.04755675196 \\
128	0.84 0.02009347867 \\
200	0.86 0.03256145745 \\
300	0.8435 0.05450531587 \\
400	0.8615 0.02960730418 \\
};

\addlegendentry{SSB-CNN, $N=2$}
\addlegendentry{SSE-CNN, $N=2$}
\addlegendentry{Z3-CNN, $\#$ filters=10}
\addlegendentry{Z3-CNN, $\#$ filters=144}

\end{axis}
\end{tikzpicture}
\caption{Performances on the synthetic dataset in terms of accuracy for a varying number of training examples. The error bars represent the CIs at $95\%$. The number of filters in the first layer for the SSB- and SSE-CNN is 2.}
\label{fig:res_lc_synth}
\end{figure}

\begin{figure}
\begin{tikzpicture}
\begin{axis}[
    axis lines = left,
    legend pos=south east,
    xmin=0, xmax=400,
    ylabel=Accuracy,
    xlabel= Number of training examples
]
\addplot [color=blue, mark=*,domain=0.7:1]
 plot [error bars/.cd, y dir = both, y explicit]
 table[row sep=crcr, x index=0, y index=1, y error index=2,]{
10  0.6655913984 0.03516374845 \\
30  0.7494623658 0.06074316736 \\
64  0.7967741937 0.03580547572  \\
128 0.8129032259 0.02470373653 \\
200 0.8279569893 0.02494833398 \\
300 0.8258064517 0.03950446429 \\
392 0.8451612903 0.01751157894 \\
};

\addplot [color=myorange, mark=*,domain=0.7:1]
 plot [error bars/.cd, y dir = both, y explicit]
 table[row sep=crcr, x index=0, y index=1, y error index=2,]{
16 0.5784946239  0.07451976428  \\
32 0.6741935488  0.06882745564  \\
64 0.7290322584  0.03911829218  \\
128 0.7376344088   0.03510974178   \\
200 0.7860215054  0.02848547569  \\
300 0.7795698925   0.04182828791  \\
400 0.7849462368 0.02337765085 \\
};

\addplot [color=mygreen, mark=*,domain=0.7:1]
 plot [error bars/.cd, y dir = both, y explicit]
 table[row sep=crcr, x index=0, y index=1, y error index=2,]{
10  0.6268817211 0.05035850889\\
30  0.6838709684 0.03679866355\\
64 0.7376344093 0.03532527321\\
128 0.7924731187 0.02994053595\\
200 0.792473119 0.03397811101\\
300 0.8215053767 0.02141179571\\
392 0.8129032263 0.01303943057\\
};

\addplot [color=myred, mark=*,domain=0.7:1]
 plot [error bars/.cd, y dir = both, y explicit]
 table[row sep=crcr, x index=0, y index=1, y error index=2,]{
16	0.5559139788 0.06566697623 \\
32	0.6462365597 0.05204846187 \\
64	0.7053763445 0.04806809908 \\
128	0.7602150544 0.02810994457 \\
200	0.7741935491 0.02782497022 \\
300	0.7967741939 0.02150023797 \\
392	0.7784946244 0.01519047688 \\
};

\addlegendentry{SSB-CNN, $N=4$}
\addlegendentry{SSE-CNN, $N=1$}
\addlegendentry{Z3-CNN, $\#$ filters=10}
\addlegendentry{Z3-CNN, $\#$ filters=144}

\end{axis}
\end{tikzpicture}
\caption{Performances on the NLST dataset in terms of accuracy for a varying number of training examples. The error bars represent the CIs at $95\%$. The number of filters in the first layer for the SSB- and SSE-CNN is 4.}
\label{fig:res_lc_NLST}
\end{figure}
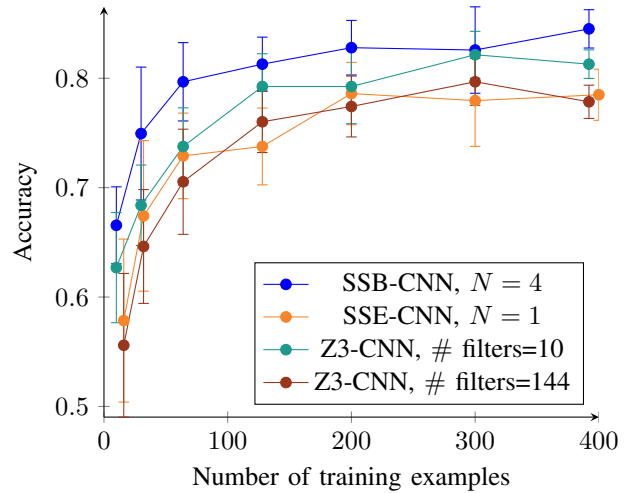

The SSB- and SSE-CNN are LRI networks and thus require neither additional training examples nor a large number of parameters to learn this property. 
In addition, they rely on compressing SH parametric representations.
For these two reasons, we expect that they will better generalize with fewer training examples (\emph{i.e.} steeper learning curve) than the standard Z3-CNN on
data for which this property is relevant. To test this hypothesis, we compare the classification performance of each method using an increasingly large number of training examples $N_s$. For the synthetic dataset, we use \mbox{$N_s = 16, 32, 64, 128, 200, 300, 400$} and for the nodule classification \mbox{$N_s= 10,30, 64,128,200,300,392$}. For each value of $N_s$, 10 repetitions are made and $N_s$ examples are randomly drawn from the same training fold as the previous experiments (Section~\ref{sec:exp_perf}).
For the SSB-CNN we report the accuracy for $N=2$ on the synthetic dataset and $N=4$ on the NLST dataset.
The accuracy of the SSE-CNN is reported for $N=2$ on the synthetic dataset and $N=1$ for the NLST dataset. 
These parameters are chosen according to the previous experiments (Section~\ref{sec:exp_perf}, Fig.~\ref{fig:res_perf_synth} and~\ref{fig:res_perf_NLST}) as they provided the best accuracy.
The experiment is also conducted with the Z3-CNN and the results are reported for both 10 and 144 filters in the convolution layer. The mean accuracy with CIs at $95\%$ of the three models and on the two datasets is reported in Fig.~\ref{fig:res_lc_synth} and~\ref{fig:res_lc_NLST}.

\vspace{-0.3cm} 

\section{Discussions}\label{sec:discussions}
\subsection{The Bispectrum is More Discriminative}
The two experiments presented in Section~\ref{sec:toyExperiment} illustrate the types of pattern information that cannot be characterized by spectral components. In these settings, the spectrum is unable to distinguish between classes that differ either by a difference of orientation between degrees (inter-degree rotation) or by intra-degree variations. This is not the case for the bispectral coefficients that allow describing functions in $L_2(\mathbb{S}^2)$ more accurately. As expected, the cost of a more complete representation is a larger number of components. However, it is possible to compute only a subset of the bispectral components depending on the task. 

In the CNN implementation of these two invariants (Section~\ref{sec:exp_perf}), we observe that the specific information captured by the SSB improves the classification performance for both datasets: as soon as the maximum degree is greater than one, the SSB-CNN outperforms the SSE-CNN (Section~\ref{sec:exp_perf}, Fig.~\ref{fig:res_perf_synth} and~\ref{fig:res_perf_NLST}).

Besides, both the SSE- and the SSB-CNN outperform the standard Z3-CNN on the synthetic data which was specifically designed to give an advantage to LRI networks. By contrast, in the nodule classification task (NLST dataset), the Z3-CNN outperforms the SSE-CNN. It seems that the simple design of the SSE-CNN fails to capture the specific signature of malignant pulmonary nodule information on these data.
However, once again, the richer invariant representation of the SSB-CNN allows outperforming even the Z3-CNN with statistical significance when $N=4$ while using approximately 22 times fewer parameters.

\vspace{-0.3cm} 

\subsection{Better Generalization of the LRI Models}
The learning curve experiment on the synthetic dataset presented in Section~\ref{sec:learningCurves} shows that both LRI designs outperform the Z3-CNN for any number of training examples.
What is more notable is the steeper learning for the two LRI networks. 
Both SSE- and SSB-CNNs seem to require the same number of training examples to reach their final performance level.
For the Z3-CNN, two networks are compared: one with 10 filters and the other with 144 filters, accounting for 7322 and 105,410 trainable parameters, respectively.
Even though the number of parameters is vastly different, the overall shape of the learning curves does not significantly change between the two Z3 networks, pointing out that the relationship between numbers of parameters and training examples is not obvious and highly depends on the architecture.

On the NLST dataset, the SSB-CNN outperforms the Z3-CNN when trained with the same number of training examples. However, the steeper learning curve of the former is less pronounced than with the synthetic dataset. We expect the gap between the two learning curves to be wider if we use deeper architecture as the difference in the number parameters will be higher.
Overall, we observe that the proposed SSB-CNN requires fewer training examples than the Z3-CNN, thanks to both the LRI property and the compressing parametric SH kernel representations.

\vspace{-0.3cm} 

\section{Conclusion}\label{sec:conclusion}

We showed that, by using the highly discriminative SSB RI descriptor, we are able to implement CNNs that are more accurate than the previously proposed SSE-CNN. Furthermore, we also observed that LRI networks can learn with fewer training examples than the more traditional Z3-CNN, which supports our hypothesis that the latter tends to misspend the parameter budget to learn data invariances and symmetries.
The main limitation of the proposed experimental evaluation is that it relies on shallow networks that would place these approaches more at the crossroad between handcrafted methods and deep learning. In future work, the LRI layers will be implemented in a deeper architecture to leverage the fewer resources that they require in comparison with a standard convolutional layer.
This is expected to constitute a major contribution to improve 3D data analysis when curated and labelled training data is scarce, which most often the case in medical image analysis.
The code is available on GitHub\footnote{\url{https://github.com/voreille/ssbcnn}, as of April 2020.}.


%

\appendices

\vspace{-0.3cm} 

\section{Clebsch-Gordan matrices}\label{app:CG}

Let us fix $n_1 , n_2 \geq 0$. 
The Clebsch-Gordan matrix $\mathrm{C}_{n_1,n_2}$ is characterized by the fact that it block-diagonalizes the Kronecker product of two Wigner-$\mathrm{D}$ matrices as
\begin{equation} \label{eq:CGandWigner}
    \mathrm{D}_{n_1}(\mathrm{R}) \otimes \mathrm{D}_{n_2}(\mathrm{R}) = \mathrm{C}_{n_1, n_2} \left[
    \bigoplus_{i=|n_1-n_2|}^{n_1+n_2} \mathrm{D}_i(\mathrm{R}) 
    \right] \mathrm{C}_{n_1, n_2}^\dag
\end{equation}
for any matrix rotation $\mathrm{R}  \in SO(3)$.
This means in particular that $\mathrm{C}_{n_1,n_2}$ has $\sum_{n= |n_1 - n_2|}^{n_1+n_2} (2n+1)$ rows and $(2n_1 +1)(2n_2+1)$ columns. These two numbers are actually equal, hence $\mathrm{C}_{n_1,n_2} \in \R^{(2n_1 +1)(2n_2+1)\times (2n_1 +1)(2n_2+1)}$, but the relation \eqref{eq:CGandWigner} also reveals the structure of the matrix, whose coefficients are indexed as $\mathrm{C}_{n_1,n_2}[ (n,m) , (m_1,m_2)]$, with $n \in \{ |n_1 - n_2| , \ldots , (n_1+n_2) \}$, $m_1 \in \{-n_1 ,\ldots , n_1 \}$, and $m_2 \in \{-n_2 ,\ldots , n_2 \}$. In the literature, the Clebsch-Gordan coefficients are often written with bracket notations, that reveal some of their symmetries~\cite{alex2011numerical}. Moreover, the Clebsch-Gordan matrix has many $0$ entries. We indeed have that
\begin{align*}
    \mathrm{C}_{n_1,n_2}[ (n,m) , (m_1,m_2)] &= 0 \text{ if } m \neq m_1 + m_2 \\
                        &= \langle n_1 m_1 n_2 m_2 | n (m_1 + m_2) \rangle,
\end{align*}
where $\langle | \rangle$ is the bracket notation used for instance in \cite[Chapter 5.3.1]{chaichian1998symmetries}.

\vspace{-0.3cm} 

\section{Proof of Proposition \ref{prop:LRI}}\label{app:LRIproof} 

The equivariance to translations is simpler and similar to the equivariance to rotations, therefore we skip it (it simply uses that $(I(\cdot - \bm{x}_0) * \kappa_n^m) (\x) = (I*\kappa_n^m)(\x - \x_0)$).
Let $\F_n(\x)$ and $\F'_n(\x)$ be the Fourier feature maps of $I$ and $I(\mathrm{R}_0 \cdot)$ respectively, with $\mathrm{R}_0 \in SO(3)$. According to \eqref{eq:steerability_ynm} applied to $\mathrm{R} = \mathrm{R}^{-1}_0$, we have that
\begin{equation} \label{eq:Ikapparota}
     \kappa_n^m (\mathrm{R}^{-1}_0 \cdot ) = \sum_{m' =-n}^n \mathrm{D}_n(\mathrm{R}_0^{-1})_{m,m'} \kappa_n^m.
\end{equation}
Moreover, we have that $(I(\mathrm{R}_0 \cdot) * \kappa_n^m ) (\x) = ( I * \kappa_n^m (\mathrm{R}^{-1}_0 \cdot) ) (\mathrm{R}_0 \x)$. Together with \eqref{eq:Ikapparota}, this implies that
\begin{equation}
    \F'_n(\x) = \left( ( I(\mathrm{R}_0 \cdot) * \kappa_n^m ) (\x) \right)_m = \F(\mathrm{R}_0 \x) \mathrm{D}_n (\mathrm{R}_0^{-1} \x). 
\end{equation}
This implies that
\begin{align*}
    \mathcal{G}_{n,n',\ell}^{\mathrm{SSB}} &\{ I ( \mathrm{R}_0 \cdot ) \} (\x) =
        \mathcal{B}\{ \F'_n(\x), \F'_{n'}(\x), \F'_{\ell}(\x) \} \\
        & = \mathcal{B}\{ \F_n(\mathrm{R}_0 \x) \mathrm{D}_n (\mathrm{R}_0^{-1}), \ldots \\
       & \qquad \F_{n'}(\mathrm{R}_0 \x) \mathrm{D}_{n'} (\mathrm{R}_0^{-1}), \F_{\ell}(\mathrm{R}_0 \x) \mathrm{D}_{\ell} (\mathrm{R}_0^{-1}) \} \\
        &= \mathcal{B}\{ \F_n(\mathrm{R}_0 \x)  , \F_{n'}(\mathrm{R}_0 \x) , \F_{\ell}(\mathrm{R}_0 \x)  \} \\
        &= \mathcal{G}_{n,n',\ell}^{\mathrm{SSB}} \{ I  \} (\mathrm{R}_0 \x),
\end{align*}
where we used the invariance of the bispectrum for the third equality. This demonstrates the equivariance of the operator $ \mathcal{G}_{n,n',\ell}^{\mathrm{SSB}}$ with respect to rotations. Finally, the locality simply follows from the fact that the convolution $I*\kappa_{n}^m (\x)$  depends on the values of $I (\x - \y)$ with $\y$ in the support of $\kappa_n^m$, which is bounded as soon as $h_n$ is compactly supported, what we assumed.

\vspace{-0.3cm}

\section*{Acknowledgment}
The authors are grateful to Michael Unser, who suggested them to consider the bispectrum as a tool to capture rotation-invariant features of 3D signals.
This work was supported by the Swiss National Science Foundation (SNSF grants 205320\_179069 and P2ELP2\_181759) and the Swiss Personalized Health Network (SPHN IMAGINE and QA4IQI projects), as well as a hardware grant from NVIDIA.

\ifCLASSOPTIONcaptionsoff
  \newpage
\fi



\bibliographystyle{IEEEtran}
\bibliography{bibtex/bib/biblio}

\begin{thebibliography}{10}
\providecommand{\url}[1]{#1}
\csname url@samestyle\endcsname
\providecommand{\newblock}{\relax}
\providecommand{\bibinfo}[2]{#2}
\providecommand{\BIBentrySTDinterwordspacing}{\spaceskip=0pt\relax}
\providecommand{\BIBentryALTinterwordstretchfactor}{4}
\providecommand{\BIBentryALTinterwordspacing}{\spaceskip=\fontdimen2\font plus
\BIBentryALTinterwordstretchfactor\fontdimen3\font minus
  \fontdimen4\font\relax}
\providecommand{\BIBforeignlanguage}[2]{{%
\expandafter\ifx\csname l@#1\endcsname\relax
\typeout{** WARNING: IEEEtran.bst: No hyphenation pattern has been}%
\typeout{** loaded for the language `#1'. Using the pattern for}%
\typeout{** the default language instead.}%
\else
\language=\csname l@#1\endcsname
\fi
#2}}
\providecommand{\BIBdecl}{\relax}
\BIBdecl

\bibitem{greenspan2016guest}
H.~Greenspan, B.~Van~Ginneken, and R.~M. Summers, ``{Guest editorial deep
  learning in medical imaging: Overview and future promise of an exciting new
  technique},'' \emph{IEEE Transactions on Medical Imaging}, vol.~35, no.~5,
  pp. 1153--1159, 2016.

\bibitem{shorten2019survey}
C.~Shorten and T.~Khoshgoftaar, ``A survey on image data augmentation for deep
  learning,'' \emph{Journal of Big Data}, vol.~6, no.~1, p.~60, 2019.

\bibitem{CoW2016b}
T.~Cohen and M.~Welling, ``Group equivariant convolutional networks,'' in
  \emph{Proceedings of The 33rd International Conference on Machine Learning},
  ser. Proceedings of Machine Learning Research, M.~F. Balcan and K.~Q.
  Weinberger, Eds., vol.~48.\hskip 1em plus 0.5em minus 0.4em\relax New York,
  New York, USA: PMLR, 20--22 Jun 2016, pp. 2990--2999.

\bibitem{weiler2017learning}
M.~Weiler, F.~A. Hamprecht, and M.~Storath, ``Learning steerable filters for
  rotation equivariant {CNN}s,'' \emph{2018 IEEE/CVF Conference on Computer
  Vision and Pattern Recognition}, pp. 849--858, 2017.

\bibitem{andrearczyk2020local}
V.~Andrearczyk, J.~Fageot, V.~Oreiller, X.~Montet, and A.~Depeursinge, ``{Local
  Rotation Invariance in 3D CNNs},'' in \emph{(submitted) Medical Image
  Analysis}, 2020.

\bibitem{eickenberg2017solid}
M.~Eickenberg, G.~Exarchakis, M.~Hirn, and S.~Mallat, ``{Solid harmonic wavelet
  scattering: Predicting quantum molecular energy from invariant descriptors of
  3D electronic densities},'' in \emph{Advances in Neural Information
  Processing Systems}, 2017, pp. 6540--6549.

\bibitem{vivaldi2006arithmetic}
F.~Vivaldi, ``The arithmetic of discretized rotations,'' in \emph{AIP
  Conference Proceedings}, vol. 826, no.~1.\hskip 1em plus 0.5em minus
  0.4em\relax American Institute of Physics, 2006, pp. 162--173.

\bibitem{ke2014rotation}
Q.~Ke and Y.~Li, ``Is rotation a nuisance in shape recognition?'' in
  \emph{Proceedings of the IEEE Conference on Computer Vision and Pattern
  Recognition}, 2014, pp. 4146--4153.

\bibitem{andrearczyk2019exploring}
V.~Andrearczyk, J.~Fageot, V.~Oreiller, X.~Montet, and A.~Depeursinge,
  ``{Exploring local rotation invariance in 3D CNNs with steerable filters},''
  in \emph{International Conference on Medical Imaging with Deep Learning},
  2019.

\bibitem{andrearczyk2019solid}
V.~Andrearczyk, V.~Oreiller, J.~Fageot, X.~Montet, and A.~Depeursinge, ``Solid
  spherical energy ({SSE}) {CNN}s for efficient 3{D} medical image analysis,''
  in \emph{Irish Machine Vision and Image Processing Conference}, 2019.

\bibitem{weiler20183d}
M.~Weiler, M.~Geiger, M.~Welling, W.~Boomsma, and T.~S. Cohen, ``3d steerable
  cnns: Learning rotationally equivariant features in volumetric data,'' in
  \emph{Advances in Neural Information Processing Systems}, 2018, pp.
  10\,381--10\,392.

\bibitem{gallier2009}
J.~Gallier, ``{Notes on spherical harmonics and linear representations of Lie
  groups},'' \url{http://www.cis.upenn.edu/~cis610/sharmonics.pdf}, 2009,
  accessed: 2020-04-13.

\bibitem{smith1997scientist}
S.~W. Smith \emph{et~al.}, \emph{The scientist and engineer's guide to digital
  signal processing}.\hskip 1em plus 0.5em minus 0.4em\relax California
  Technical Pub. San Diego, 1997.

\bibitem{kakarala2010}
R.~Kakarala and D.~Mao, ``A theory of phase-sensitive rotation invariance with
  spherical harmonic and moment-based representations,'' in \emph{Computer
  Vision and Pattern Recognition (CVPR), 2010 IEEE Conference on}.\hskip 1em
  plus 0.5em minus 0.4em\relax IEEE, 2010, pp. 105--112.

\bibitem{depeursinge2018rotation}
A.~Depeursinge, J.~Fageot, V.~Andrearczyk, J.~P. Ward, and M.~Unser, ``Rotation
  invariance and directional sensitivity: {S}pherical harmonics versus
  radiomics features,'' in \emph{International Workshop on Machine Learning in
  Medical Imaging}.\hskip 1em plus 0.5em minus 0.4em\relax Springer, 2018, pp.
  107--115.

\bibitem{OPM2002}
T.~Ojala, M.~Pietik{\"{a}}inen, and T.~M{\"{a}}enp{\"{a}}{\"{a}},
  ``{Multiresolution gray--scale and rotation invariant texture classification
  with local binary patterns},'' \emph{IEEE Transactions on Pattern Analysis
  and Machine Intelligence}, vol.~24, no.~7, pp. 971--987, July 2002.

\bibitem{VaZ2005}
M.~Varma and A.~Zisserman, ``A statistical approach to texture classification
  from single images,'' \emph{International Journal of Computer Vision},
  vol.~62, no. 1-2, pp. 61--81, 2005.

\bibitem{freeman1991design}
W.~Freeman and E.~Adelson, ``The design and use of steerable filters,''
  \emph{IEEE Transactions on Pattern Analysis \& Machine Intelligence}, no.~9,
  pp. 891--906, 1991.

\bibitem{Unser2013steerable}
M.~Unser and N.~Chenouard, ``A unifying parametric framework for {2D} steerable
  wavelet transforms,'' \emph{SIAM Journal on Imaging Sciences}, vol.~6, no.~1,
  pp. 102--135, 2013.

\bibitem{perona1992steerable}
P.~Perona, ``Steerable-scalable kernels for edge detection and junction
  analysis,'' in \emph{European Conference on Computer Vision}.\hskip 1em plus
  0.5em minus 0.4em\relax Springer, 1992, pp. 3--18.

\bibitem{DicenteCid2017}
Y.~{Dicente Cid}, H.~M{\"{u}}ller, A.~Platon, P.~Poletti, and A.~Depeursinge,
  ``3{D} solid texture classification using locally-oriented wavelet
  transforms,'' \emph{IEEE Transactions on Image Processing}, vol.~26, pp.
  1899--1910, 2017.

\bibitem{fageot2018principled}
J.~Fageot, V.~Uhlmann, Z.~P{\"u}sp{\"o}ki, B.~Beck, M.~Unser, and
  A.~Depeursinge, ``Principled design and implementation of steerable
  detectors,'' \emph{arXiv preprint arXiv:1811.00863}, 2018.

\bibitem{DPW2017}
A.~Depeursinge, Z.~P\"{u}sp\"{o}ki, J.~P. Ward, and M.~Unser, ``{Steerable
  Wavelet Machines (SWM): Learning Moving Frames for Texture Classification},''
  \emph{IEEE Transactions on Image Processing}, vol.~26, no.~4, pp. 1626--1636,
  2017.

\bibitem{flusser2009moments}
J.~Flusser, B.~Zitova, and T.~Suk, \emph{{Moments and moment invariants in
  pattern recognition}}.\hskip 1em plus 0.5em minus 0.4em\relax John Wiley \&
  Sons, 2009.

\bibitem{kakarala2012bispectrum}
R.~Kakarala, ``The bispectrum as a source of phase-sensitive invariants for
  fourier descriptors: a group-theoretic approach,'' \emph{Journal of
  Mathematical Imaging and Vision}, vol.~44, no.~3, pp. 341--353, 2012.

\bibitem{zucchelli2020computational}
M.~Zucchelli, S.~Deslauriers-Gauthier, and R.~Deriche, ``{A computational
  Framework for generating rotation invariant features and its application in
  diffusion MRI},'' \emph{Medical Image Analysis}, vol.~60, p. 101597, 2020.

\bibitem{kondor2018generalization}
R.~Kondor and S.~Trivedi, ``On the generalization of equivariance and
  convolution in neural networks to the action of compact groups,'' \emph{arXiv
  preprint arXiv:1802.03690}, 2018.

\bibitem{cohen2019general}
T.~S. Cohen, M.~Geiger, and M.~Weiler, ``A general theory of equivariant {CNN}s
  on homogeneous spaces,'' in \emph{Advances in Neural Information Processing
  Systems}, 2019, pp. 9142--9153.

\bibitem{winkels2019pulmonary}
M.~Winkels and T.~S. Cohen, ``Pulmonary nodule detection in {CT} scans with
  equivariant {CNN}s,'' \emph{Medical image analysis}, vol.~55, pp. 15--26,
  2019.

\bibitem{worrall2018cubenet}
D.~Worrall and G.~Brostow, ``{CubeNet}: {E}quivariance to {3D} rotation and
  translation,'' \emph{ECCV, Lecture Notes in Computer Science}, vol. 11209,
  pp. 585--602, 2018.

\bibitem{andrearczyk2018rotational}
V.~Andrearczyk and A.~Depeursinge, ``Rotational 3{D} texture classification
  using group equivariant {CNN}s,'' \emph{arXiv:1810.06889}, 2018.

\bibitem{bekkers2018roto}
E.~J. Bekkers, M.~W. Lafarge, M.~Veta, K.~A. Eppenhof, J.~P. Pluim, and
  R.~Duits, ``Roto-translation covariant convolutional networks for medical
  image analysis,'' in \emph{International Conference on Medical Image
  Computing and Computer-Assisted Intervention}.\hskip 1em plus 0.5em minus
  0.4em\relax Springer, 2018, pp. 440--448.

\bibitem{cohen2016steerable}
T.~Cohen and M.~Welling, ``Steerable {CNN}s,'' \emph{arXiv:1612.08498}, 2016.

\bibitem{coxeter1961introduction}
H.~S.~M. Coxeter, \emph{Introduction to geometry}.\hskip 1em plus 0.5em minus
  0.4em\relax New York, London, 1961.

\bibitem{WGT2016}
D.~E. Worrall, S.~J. Garbin, D.~Turmukhambetov, and G.~J. Brostow, ``Harmonic
  networks: Deep translation and rotation equivariance,'' \emph{2017 IEEE
  Conference on Computer Vision and Pattern Recognition (CVPR)}, pp.
  7168--7177, 2016.

\bibitem{kondor2018clebsch}
R.~Kondor, Z.~Lin, and S.~Trivedi, ``Clebsch-{G}ordan nets: a fully {F}ourier
  space spherical convolutional neural network,'' in \emph{Advances in Neural
  Information Processing Systems}, 2018, pp. 10\,117--10\,126.

\bibitem{cohen2018spherical}
T.~Cohen, M.~Geiger, J.~K{\"o}hler, and M.~Welling, ``Spherical {CNN}s,''
  \emph{arXiv preprint arXiv:1801.10130}, 2018.

\bibitem{bekkers2019b}
E.~J. Bekkers, ``{B-Spline CNNs on Lie groups},'' \emph{arXiv preprint
  arXiv:1909.12057}, 2019.

\bibitem{varshalovich1988quantum}
D.~Varshalovich, A.~Moskalev, and V.~Khersonskii, \emph{Quantum theory of
  angular momentum}.\hskip 1em plus 0.5em minus 0.4em\relax World Scientific,
  1988.

\bibitem{kakarala2011viewpoint}
R.~Kakarala, P.~Kaliamoorthi, and W.~Li, ``Viewpoint invariants from
  three-dimensional data: the role of reflection in human activity
  understanding,'' in \emph{CVPR 2011 WORKSHOPS}.\hskip 1em plus 0.5em minus
  0.4em\relax IEEE, 2011, pp. 57--62.

\bibitem{portilla2000parametric}
J.~Portilla and E.~P. Simoncelli, ``A parametric texture model based on joint
  statistics of complex wavelet coefficients,'' \emph{International journal of
  computer vision}, vol.~40, no.~1, pp. 49--70, 2000.

\bibitem{AnW2016}
V.~Andrearczyk and P.~Whelan, ``Using filter banks in convolutional neural
  networks for texture classification,'' \emph{Pattern Recognition Letters},
  vol.~84, pp. 63--69, 2016.

\bibitem{he2015delving}
K.~He, X.~Zhang, S.~Ren, and J.~Sun, ``Delving deep into rectifiers: Surpassing
  human-level performance on imagenet classification,'' in \emph{Proceedings of
  the IEEE international conference on computer vision}, 2015, pp. 1026--1034.

\bibitem{alex2011numerical}
A.~Alex, M.~Kalus, A.~Huckleberry, and J.~von Delft, ``A numerical algorithm
  for the explicit calculation of $su(n)$ and $sl(n, c)$ {C}lebsch--{G}ordan
  coefficients,'' \emph{Journal of Mathematical Physics}, vol.~52, no.~2, p.
  023507, 2011.

\bibitem{chaichian1998symmetries}
M.~Chaichian and R.~Hagedorn, \emph{Symmetries in quantum mechanics: from
  angular momentum to supersymmetry}, 1st~ed.\hskip 1em plus 0.5em minus
  0.4em\relax CRC Press, 1998.

\end{thebibliography}

\end{document}